\documentclass{nle}

\usepackage{amssymb}
\usepackage[T1]{fontenc}
\usepackage[table]{xcolor}
\usepackage{enumitem}
\usepackage{multirow}
\usepackage{subcaption}
\usepackage{pifont}
\usepackage{hyperref}

\usepackage{booktabs, booktabs, multirow}

\usepackage{graphicx}
\usepackage{soul}
\usepackage{enumitem}
\usepackage{tcolorbox}
\usepackage{mdframed}
\usepackage{xcolor}

\usepackage{mdframed}

\usepackage{todonotes}
 \usepackage{lipsum}

 \newcommand{\data}{\textsc{Unit}}
  \newcommand{\model}{GPT-2$^{\textsc{U}}$}

\usepackage{amsmath}
\usepackage{graphicx}
\usepackage{natbib}
\newcommand\blfootnote[1]{%
  \begingroup
  \renewcommand\thefootnote{}\footnote{#1}%
  \addtocounter{footnote}{-1}%
  \endgroup
}

\ifpdf%
\usepackage{epstopdf}%
\else%
\fi
\usepackage{multirow}

\begin{document}
\label{firstpage}

\lefttitle{\LaTeX\ Supplement}
\righttitle{Natural Language Engineering}

\papertitle{Article}

\jnlPage{1}{00}
\jnlDoiYr{2019}
\doival{10.1017/xxxxx}

\title{Dialogue Agents 101: A Beginner's Guide to Critical Ingredients for Designing Effective Conversational Systems}

\begin{authgrp}
\author{Shivani Kumar}
\affiliation{Indraprastha Institute of Information Technology, Delhi; \email{shivaniku@iiitd.ac.in}}
\author{Sumit Bhatia}
\affiliation{Media and Data Science Research Lab, Adobe; \email{sumit.bhatia@adobe.com}}
\author{Milan Aggarwal}
\affiliation{Media and Data Science Research Lab, Adobe; \email{milaggar@adobe.com}}
\author{ Tanmoy Chakraborty}
\affiliation{Indian Institute of Technology, Delhi; \email{tanchak@iitd.ac.in}}
\end{authgrp}

\history{(Received xx xxx xxx; revised xx xxx xxx; accepted xx xxx xxx)}

\begin{abstract}
Sharing ideas through communication with peers is the primary mode of human interaction. Consequently, extensive research has been conducted in the area of conversational AI, leading to an increase in the availability and diversity of conversational tasks, datasets, and methods. However, with numerous tasks being explored simultaneously, the current landscape of conversational AI has become fragmented. Consequently, initiating a well-thought-out model for a dialogue agent can pose significant challenges for a practitioner. Towards highlighting the critical ingredients needed for a practitioner to design a dialogue agent from scratch, the current study provides a comprehensive overview of the primary characteristics of a dialogue agent, the supporting tasks, their corresponding open-domain datasets, and the methods used to benchmark these datasets. We observe that different methods have been used to tackle distinct dialogue tasks. However, building separate models for each task is costly and does not leverage the correlation among the several tasks of a dialogue agent. As a result, recent trends suggest a shift towards building unified foundation models. To this end, we propose \data, a \textsc{Un}ified d\textsc{i}alogue datase\textsc{t} constructed from conversations of varying datasets for different dialogue tasks capturing the nuances for each of them.
We then train a Unified dialogue foundation model, \model\ and present a concise comparative performance of \model\ against existing large language models. We also examine the evaluation strategies used to measure the performance of dialogue agents and highlight the scope for future research in the area of conversational AI with a thorough discussion of popular models such as ChatGPT.
\end{abstract}

\maketitle

\section{Introduction} 
The\blfootnote{\noindent\textbf{Competing interests:} Shivani Kumar is pursuing her PhD at Indraprastha Institute of Information Technology Delhi. Sumit Bhatia and Milan Aggarwal are employed at Adobe. Tanmoy Chakraborty is employed at Indian Institute of Technology Delhi.} significance of conversations as the fundamental medium of interaction transcends cultural boundaries~\citep{dingemanse2014conversation}.  Consequently, interacting with machines and seeking information via conversational interfaces is an instinctive and familiar way for humans~\citep{cis-zamani} as evidenced by the success of dialogue systems such as Apple's SIRI\footnote{\url{https://www.apple.com/in/siri/}}, Amazon's Alexa\footnote{\url{https://alexa.amazon.com/}}, and most recently, ChatGPT\footnote{\url{https://openai.com/blog/chatgpt}}. Moreover, dialogue-based systems\footnote{We use dialogue-based systems, chatbots, conversational systems, and dialogue agents interchangeably in this article.} have extensively been used for customer support~\citep{botea2019generating,feigenblat2021tweetsumm}, mental health support~\citep{kretzschmar2019can}, and counseling~\citep{ 10.1145/3488560.3498509, tewari2021survey}.

Designing practical dialogue-based systems, however, is a challenging endeavour as there are important questions that one needs to answer before embarking on developing such a system. Critical considerations include determining the types of queries the system should anticipate (e.g., chit-chat versus informational), deciding whether to incorporate an external knowledge source, and determining the level of natural language understanding the system should support. Previous surveys in the field of dialogue-based systems have predominantly focused on examining specific system components or narrow subsets of tasks and techniques. For instance, recent surveys have delved into areas such as dialogue summarization~\citep{tuggener-etal-2021-summarizing,feng2022survey}, text-to-SQL~\citep{qin2022survey}, question answering~\citep{pandya2021question}, dialogue management using deep learning~\citep{10.1145/3166054.3166058} and reinforcement learning ~\citep{dai2021survey}.

While the surveys noted above provide comprehensive insights into their respective domains, this abundance of information can make it overwhelming for both novice and experienced researchers and professionals to identify the essential components required for building their dialogue-based systems. In contrast, we adopt a broader perspective and offer a panoramic view of the various constituents comprising a dialogue-based system, elucidate the individual tasks involved in their development, and highlight the typical datasets and state-of-the-art methodologies employed for designing and evaluating these components.
{Consequently, the title `Dialogue Agents 101' is a deliberate choice aiming to convey that the article serves as an introductory guide or primer to the fundamental concepts and principles associated with dialogue agents. In academic settings, `101' is often used to denote introductory or basic-level courses, and here, it suggests that the article provides foundational knowledge for readers who may be new to the topic of dialogue agents.}
With this comprehensive survey, we aspire to assist beginners and practitioners in making well-informed decisions while developing systems for their applications.
{Our specific objective is to comprehensively encompass all \textbf{prominent open-source textual English} dialogue datasets across major dialogue tasks. That is, every dataset under consideration in our study meets four conditions: (i) it must be widely recognized within its respective field; ii) it should incorporate a textual component in both input and output; (iii) it must be publicly accessible, and; and (iv) it must be designed for English.}

To identify relevant material for our survey, we conducted a thorough search of the {\tt Papers With Code} website\footnote{\url{https://paperswithcode.com/}}
to identify all relevant tasks and datasets related to dialogue agents. Our goal was to gather and systematically organize different types of tasks that may be required for developing various dialogue-agents; and understand the methods for performing these tasks,  and datasets that are typically used to train and evaluate models for these tasks.
{From the initial list obtained from 
{\tt Papers With Code}, we then queried {\tt Google Scholar} for publications and followed the citation threads to gather relevant literature for each task, encompassing datasets and articles proposed well before the establishment of the platforms.
{We emphasize that while {\tt Papers With Code} functioned as our reference for locating pertinent literature, its principal values lay in pinpointing the key problem statements investigated within the domain of dialogue agents.}}

{While delving into contemporary deep learning methods in this investigation, it is crucial to acknowledge the rich history of research in dialogue agents. Long before the advent of deep learning, researchers were actively engaged in developing computational methods to facilitate meaningful interactions between machines and humans \citep{bayer-etal-2001-dialogue, 10.1145/365153.365168}. In the nascent stages of dialogue agent development, researchers heavily relied on rule-based systems \citep{webb2000rule, McTear2021}. Human experts meticulously crafted these systems, incorporating predefined rules and decision trees to interpret user inputs and generate appropriate responses. Classification tasks, such as intent detection and slot filling, often involved rule-based pattern matching \citep{5697471, ren2018joint} and template-based approaches \citep{onyshkevych1993template, mcroy2003augmented} to identify the user's intention based on specific keywords or syntactic structures. Generative tasks, such as response generation, posed a significant challenge without deep learning techniques. Early approaches leveraged handcrafted templates \citep{chu1998collaborative, 10.1145/365153.365168}, where responses were generated by combining predefined phrases or sentences. This method, however, lacked the flexibility to generate contextually relevant and nuanced responses, hindering the natural flow of conversations.

As computational capabilities advanced, statistical methods started gaining traction in dialogue agent development. Hidden Markov Models (HMMs) \citep{1165342} and finite-state machines \citep{fsm-inbook} were applied to model the probabilistic nature of language and user interactions \citep{williams2003probabilistic, williams2005factored}. These models enabled a more dynamic and probabilistic approach to intent detection and slot filling, contributing to the improvement of dialogue system performance \citep{1175148, zhao-etal-2004-discriminative}. From rule-based systems and template-based approaches to early statistical models, researchers laid the groundwork for the sophisticated deep learning methodologies that dominate the contemporary landscape we aim to study in this survey.} To summarize, our key contributions are as follows.

\begin{enumerate}[noitemsep,leftmargin=*]
    \item We propose an \textbf{in-depth taxonomy} for different components and modules involved in building a dialogue agent (Figure~\ref{fig:taxonomy}). We take a practitioner's view point and develop the taxonomy in terms of features of the underlying system and discuss at length the role played by each of the features in the overall system (Section~\ref{sec:dialogue_agents}).
    
    \item  Next, we present a comprehensive overview of different tasks and datsets in the literature and relate them to the features as identified in the proposed taxonomy (Table~\ref{tab:tasks_taxonomy}). We identify eleven broad categories of tasks related to dialogue-based systems and present a detailed overview of different methods for each task and datasets used for evaluating these tasks (Section~\ref{sec:tasks}). Our goal is to help the reader identify key techniques and datasets available for the tasks relevant to their applications.
    
    \item We present \data\footnote{We make \data\ public on \url{https://github.com/LCS2-IIITD/UNIT.git}}, 
    a large scale \textbf{unified dialogue dataset}, consisting of more than 4.8M dialogues and 441M tokens, which combine the various dialogue datasets described in Section~\ref{sec:standardised_data}. {Since \data\ is made from the dialogues of open-sourced datasets, it is free to use for any research purposes.} This effort is motivated by the recent trends suggesting a shift towards building unified foundation models~\citep{zhou2023comprehensive} that are pre-trained on large datasets and generalize to a variety of tasks. We make \data\ available to the research community with a goal to spark research efforts towards development of foundation models optimized for dialogues. We use \data\ to further pretrain popular open dialogue foundation models and show how it can help improving their performance on various dialogue tasks (Section~\ref{sec:experiments}).
\end{enumerate}

\section{Designing a Dialogue Agent}
\label{sec:dialogue_agents}
\begin{figure*}[h!]
    \centering
    \includegraphics[width=\textwidth]{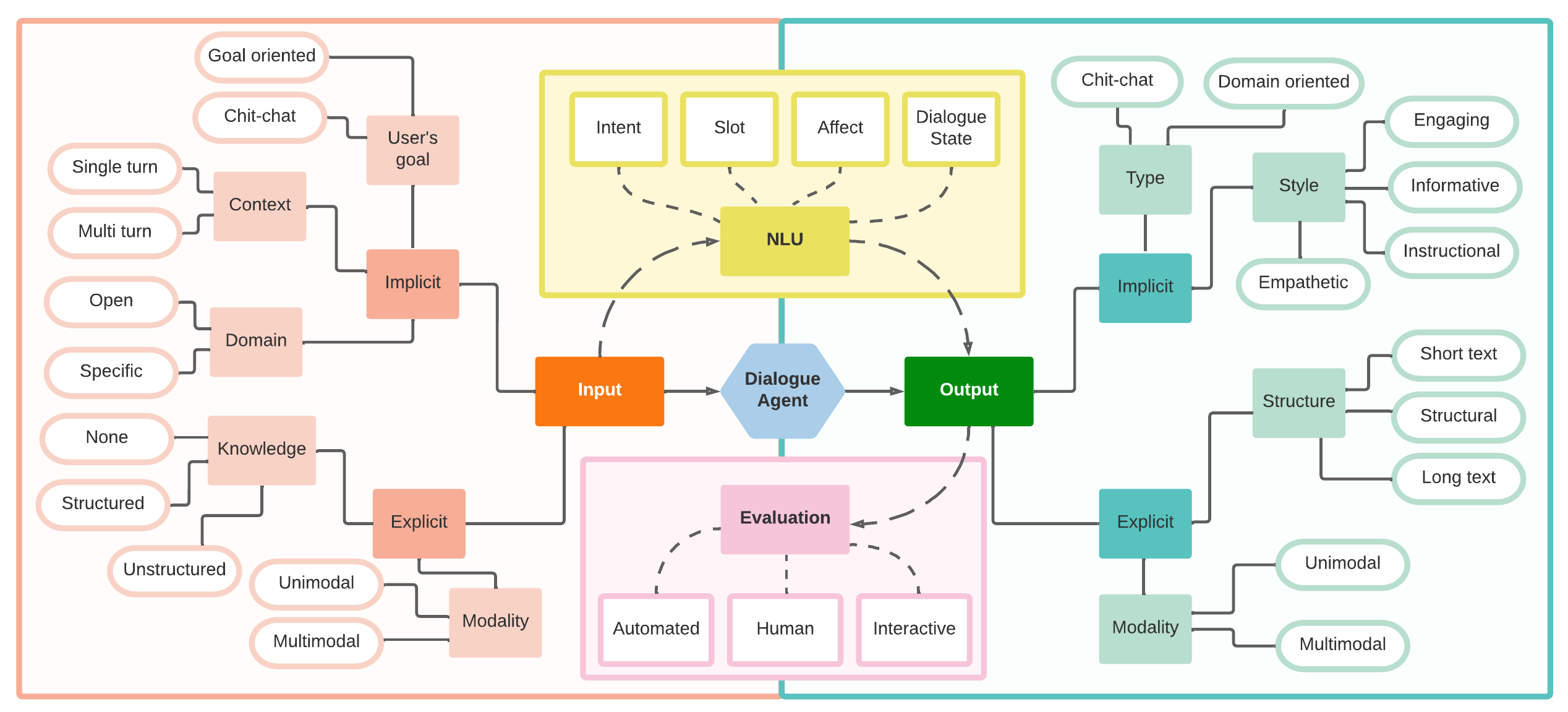}
    \caption{{\bf A taxonomic overview of a dialogue agent.} The major components for designing a complete pipeline of a dialogue agent are -- input(s), natural language understanding (NLU), generated output(s), and model evaluation. Each component can be further divided based on the characteristics required in the final dialogue agent.}
    \label{fig:taxonomy}
\end{figure*}

\begin{figure}
    \centering
    \resizebox{0.5\textwidth}{!}{%
    \subfloat[Goal-oriented single-turn dialogue of a single domain with structured knowledge and multimodal input]{
        \includegraphics[width=0.4\textwidth]{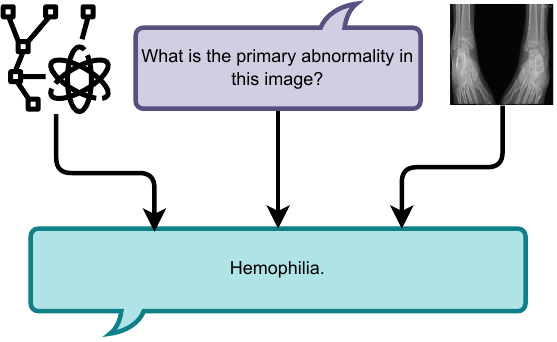}
    }} \hfill 
    \resizebox{0.35\textwidth}{!}{%
    \subfloat[Chit-chat multi turn dialogue of a open domain with no external knowledge and unimodal input]{
        \includegraphics[width=0.31\textwidth]{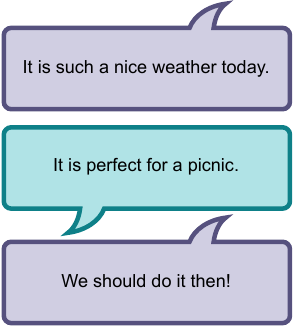}
    }}
    \caption{Dialogues highlighting different attributes of a dialogue agent input and output.}
    \label{fig:input_attr}
\end{figure}

\noindent Before developing a dialogue agent, several crucial decisions must be made to determine the appropriate architecture for the agent. Figure \ref{fig:taxonomy} illustrates a comprehensive overview of these decisions, which provides a taxonomic framework for structuring the development process. A clear understanding of the end goal we aim to achieve from a dialogue agent is crucial for effective communication \citep{pomerantz2011conversation}. 
For instance, questions such as ``Do we want the dialogue agent to carry out goal-oriented or chit-chat conversations?'' and {``Does the agent need any external knowledge to answer user queries?'' should be answered.}
Figure \ref{fig:input_attr} highlights the different type of dialogues based on the different attributes of the input and output of the sysetem as discussed below.

\subsection{Input to the System } 
After establishing the end-goal of our dialogue agent, 
it is essential to determine the various factors that will inform the input to the agent \citep{8536470}.
Our contention is that the input can possess both implicit and explicit properties, depending on the task at hand. 

\paragraph{Implicit Attributes}
We classify the characteristics of the input which are not explicitly apparent from the input as implicit attributes of the input. This inherent information can be decided based on three aspects -- the user's goal \citep{muise2019planning}, the domain of the dialogues \citep{budzianowski-etal-2018-multiwoz}, and the context needed to carry out the end task \citep{kiela2019makes}. Depending on the objective of the dialogue agent, the user could want to achieve some goal, such as making a restaurant reservation, booking an airline ticket, or resolving technical queries. For such goal-oriented dialogue agents, the input from the user is expected to differ from that received for general chit-chat \citep{muise2019planning}. Goal-oriented dialogue agents are often designed to operate within a particular domain, while chit-chat-based agents are more versatile and are expected to handle a broader range of conversations \citep{zhang-etal-2018-personalizing}. In addition to the user's goal and the agent's domain, the conversation context also plays a crucial role in achieving the agent's objective \citep{kiela2019makes}. For example, utterance-level intent detection may not require understanding deep conversation context, while summarizing dialogues would require a complete understanding of the context \citep{gliwa-etal-2019-samsum}.

\paragraph{Explicit Attributes} Apart from the implicit aspects of the dialogue agent's input, various input characteristics are external in nature and should be considered while building a dialogue agent. These aspects constitute the input modality \citep{doi:10.1080/10350330.2018.1504732} and any additional knowledge supplied to the agent \citep{dinan2018wizard}. Input can be unimodal, such as text or audio, or in a combination of modalities, such as an image and associated text, as in the case of visual question-answering systems \citep{Parvaneh2019ShowPA}. Furthermore, additional knowledge may be required to generate appropriate responses. For example, in a chit-chat setting, the agent may need to possess commonsense knowledge \citep{strathearn-gkatzia-2022-task2dial}, while in a question-answering setting, the agent may need to access relevant documents to provide accurate responses \citep{feng-etal-2020-doc2dial}. Therefore, any explicit knowledge supplied to the dialogue agent can be structured, like a tree or a tuple, or unstructured, like a document.

\subsection{Natural Language Understanding}
After receiving input from the user, the subsequent step involves comprehension \citep{liu2021benchmarking}. Regardless of whether the task is domain-specific or open-domain, specific attributes of the input must be identified to determine the required output. We identify four primary attributes that need to be identified from the input text -- the user's {\bf intent} \citep{casanueva-etal-2020-efficient}, any {\bf slots} needed to fulfill the intent \citep{10.1145/3547138}, {\bf affective} understanding of the input \citep{ruusuvuori2012emotion}, and the {\bf dialogue state} of the input utterance \citep{balaraman-etal-2021-recent}. While intent and slots are directly useful for a domain-specific agent to effectively complete a task, affect understanding and dialogue state tracking is also critical for a chit-chat-based agent. Affect understanding involves comprehending the user's emotion \citep{poria-etal-2019-meld}, sarcasm \citep{castro-etal-2019-towards}, and amusement \citep{Bedi2021MultimodalSD} in the input utterance. Furthermore, dialogue state tracking checks the type of utterance received by the agent, such as question, clarification, or guidance. Understanding these aspects is essential to determine the utterance's underlying meaning and provide relevant responses for the task.

\subsection{Output of the System}
The output generated by the dialogue agent, akin to its input, possesses both implicit and explicit attributes, described below. 

\paragraph{Implicit Attributes} Implicit attributes refer to the output's type \citep{Rastogi_Zang_Sunkara_Gupta_Khaitan_2020} and style \citep{su2020stylistic, troiano_velutharambath_klinger_2023}, while explicit attributes pertain to its modality \citep{sun-etal-2022-multimodal} and structure \citep{yu-etal-2018-spider}. Congruent to the user's goal in the input scenario, the type of attribute should be decided based on the end task needed to be performed by the dialogue agent. 
Depending on the end task of the agent, the resulting output can be informative \citep{feng-etal-2020-doc2dial}, engaging \citep{zhang-etal-2018-personalizing}, instructional \citep{strathearn-gkatzia-2022-task2dial}, or empathetic \citep{rashkin-etal-2019-towards}. For instance, a question-answering-based bot should be informative, while a cooking recipe bot should be more instructional. Both bots need not be empathetic in nature.

\paragraph{Explicit Attributes} While the inherent properties of the output text are critical to assess, the explicit attributes, such as modality and structure, must be considered before finalizing the dialogue agent's architecture. Modality decides whether the required output is unimodal (such as text) or multimodal (such as text with an image). Moreover, the output can be structured differently based on the task at hand. 
For instance, tasks such as text-to-SQL \citep{yu-etal-2018-spider} conversion require the output to adhere to a certain structure.
After considering various aspects of the input, output, and understanding based on the end task,
the generated output is evaluated to gauge the performance of the resultant dialogue agent \citep{deriu2021survey}. A detailed discussion about the evaluation can be found in Section \ref{sec:evaluation}.

\section{Tasks, Datasets and Methods}
\label{sec:tasks}
\begin{table*}[!t]
\centering
\resizebox{\textwidth}{!}{%
\begin{tabular}{l|l|l|l|lllllllllll|lllllllllll|l}
\toprule
\multicolumn{2}{l|}{\multirow{4}{*}{\textbf{Type}}} & \multicolumn{1}{c|}{\multirow{4}{*}{\textbf{Task}}} & \multicolumn{1}{c|}{\multirow{4}{*}{\textbf{Datasets}}} & \multicolumn{11}{c|}{\textbf{Input}} & \multicolumn{11}{c|}{\textbf{Output}} & \multicolumn{1}{l}{\multirow{4}{*}{\textbf{Size}}} \\ \cmidrule{5-26} 

\multicolumn{2}{c|}{} & \multicolumn{1}{c|}{} & \multicolumn{1}{c|}{} & \multicolumn{6}{c|}{\textbf{Implicit}} & \multicolumn{5}{c|}{\textbf{Explicit}} & \multicolumn{6}{c|}{\textbf{Implicit}} & \multicolumn{5}{c}{\textbf{Explicit}} \\ \cmidrule{5-26} 

\multicolumn{2}{c|}{} & \multicolumn{1}{c|}{} & \multicolumn{1}{c|}{} & \multicolumn{2}{c|}{\textbf{User's goal}} & \multicolumn{2}{c|}{\textbf{Domain}} & \multicolumn{2}{c|}{\textbf{Context}} & \multicolumn{2}{c|}{\textbf{Modality}} & \multicolumn{3}{c|}{\textbf{Knowledge}} & \multicolumn{2}{c|}{\textbf{Type}} & \multicolumn{4}{c|}{\textbf{Style}} & \multicolumn{2}{c|}{\textbf{Modality}} & \multicolumn{3}{c}{\textbf{Structure}} \\ \cmidrule{5-26} 

\multicolumn{2}{c|}{} & \multicolumn{1}{c|}{} & \multicolumn{1}{c|}{} & \multicolumn{1}{c|}{\rotatebox{90}{\textbf{CC}}} & \multicolumn{1}{c|}{\rotatebox{90}{\textbf{GO}}} & \multicolumn{1}{c|}{\rotatebox{90}{\textbf{Open}}} & \multicolumn{1}{c|}{\rotatebox{90}{\textbf{Spc}}} & \multicolumn{1}{c|}{\rotatebox{90}{\textbf{ST}}} & \multicolumn{1}{c|}{\rotatebox{90}{\textbf{MT}}} & \multicolumn{1}{c|}{\rotatebox{90}{\textbf{U}}} & \multicolumn{1}{c|}{\rotatebox{90}{\textbf{M}}} & \multicolumn{1}{c|}{\rotatebox{90}{\textbf{None}}} & \multicolumn{1}{c|}{\rotatebox{90}{\textbf{Unstr}}} & \multicolumn{1}{c|}{\rotatebox{90}{\textbf{Str}}} & \multicolumn{1}{c|}{\rotatebox{90}{\textbf{CC}}} & \multicolumn{1}{c|}{\rotatebox{90}{\textbf{GO}}} & \multicolumn{1}{c|}{\rotatebox{90}{\textbf{Eng}}} & \multicolumn{1}{c|}{\rotatebox{90}{\textbf{Inf}}} & \multicolumn{1}{c|}{\rotatebox{90}{\textbf{Instr}}} & \multicolumn{1}{c|}{\rotatebox{90}{\textbf{Emp}}} & \multicolumn{1}{c|}{\rotatebox{90}{\textbf{U}}} & \multicolumn{1}{c|}{\rotatebox{90}{\textbf{M}}} & \multicolumn{1}{c|}{\rotatebox{90}{\textbf{Shor}}} & \multicolumn{1}{c|}{\rotatebox{90}{\textbf{Long}}} & \multicolumn{1}{c}{\rotatebox{90}{\textbf{Struct}}} \\ \midrule

\multicolumn{1}{l|}{\multirow{29}{*}{\textbf{\rotatebox{90}{Generative}}}} & \multirow{6}{*}{\textbf{\rotatebox{90}{Transformation}}} & DR & CANARD {{\citep{elgohary-etal-2019-unpack}}} & \multicolumn{1}{l|}{\checkmark} & \multicolumn{1}{l|}{-} & \multicolumn{1}{l|}{\checkmark} & \multicolumn{1}{l|}{-} & \multicolumn{1}{l|}{-} & \multicolumn{1}{l|}{\checkmark} & \multicolumn{1}{l|}{\checkmark} & \multicolumn{1}{l|}{-} & \multicolumn{1}{l|}{\checkmark} & \multicolumn{1}{l|}{-} & - & \multicolumn{1}{l|}{\checkmark} & \multicolumn{1}{l|}{-} & \multicolumn{1}{l|}{\checkmark} & \multicolumn{1}{l|}{\checkmark} & \multicolumn{1}{l|}{-} & \multicolumn{1}{l|}{-} & \multicolumn{1}{l|}{\checkmark} & \multicolumn{1}{l|}{-} & \multicolumn{1}{l|}{-} & \multicolumn{1}{l|}{\checkmark} & - & {40} \\ \cmidrule{3-27} 

\multicolumn{1}{l|}{} &  & \multirow{2}{*}{DS} & DialogSum {\citep{chen-etal-2021-dialogsum}} & \multicolumn{1}{l|}{\checkmark} & \multicolumn{1}{l|}{-} & \multicolumn{1}{l|}{\checkmark} & \multicolumn{1}{l|}{-} & \multicolumn{1}{l|}{-} & \multicolumn{1}{l|}{\checkmark} & \multicolumn{1}{l|}{\checkmark} & \multicolumn{1}{l|}{-} & \multicolumn{1}{l|}{\checkmark} & \multicolumn{1}{l|}{-} & - & \multicolumn{1}{l|}{\checkmark} & \multicolumn{1}{l|}{-} & \multicolumn{1}{l|}{\checkmark} & \multicolumn{1}{l|}{-} & \multicolumn{1}{l|}{-} & \multicolumn{1}{l|}{-} & \multicolumn{1}{l|}{\checkmark} & \multicolumn{1}{l|}{-} & \multicolumn{1}{l|}{-} & \multicolumn{1}{l|}{\checkmark} & - & {13} \\ 

\multicolumn{1}{l|}{} &  & & SAMSum Corpus {\citep{gliwa-etal-2019-samsum}} & \multicolumn{1}{l|}{\checkmark} & \multicolumn{1}{l|}{-} & \multicolumn{1}{l|}{\checkmark} & \multicolumn{1}{l|}{-} & \multicolumn{1}{l|}{-} & \multicolumn{1}{l|}{\checkmark} & \multicolumn{1}{l|}{\checkmark} & \multicolumn{1}{l|}{-} & \multicolumn{1}{l|}{\checkmark} & \multicolumn{1}{l|}{-} & - & \multicolumn{1}{l|}{\checkmark} & \multicolumn{1}{l|}{-} & \multicolumn{1}{l|}{\checkmark} & \multicolumn{1}{l|}{-} & \multicolumn{1}{l|}{-} & \multicolumn{1}{l|}{-} & \multicolumn{1}{l|}{\checkmark} & \multicolumn{1}{l|}{-} & \multicolumn{1}{l|}{-} & \multicolumn{1}{l|}{\checkmark} & - & {16} \\ \cmidrule{3-27} 

\multicolumn{1}{l|}{} &  & \multirow{3}{*}{D2S} & CoSQL {\citep{yu-etal-2019-cosql}} & \multicolumn{1}{l|}{-} & \multicolumn{1}{l|}{\checkmark} & \multicolumn{1}{l|}{-} & \multicolumn{1}{l|}{\checkmark} & \multicolumn{1}{l|}{\checkmark} & \multicolumn{1}{l|}{-} & \multicolumn{1}{l|}{\checkmark} & \multicolumn{1}{l|}{-} & \multicolumn{1}{l|}{-} & \multicolumn{1}{l|}{-} & \checkmark & \multicolumn{1}{l|}{-} & \multicolumn{1}{l|}{\checkmark} & \multicolumn{1}{l|}{-} & \multicolumn{1}{l|}{-} & \multicolumn{1}{l|}{\checkmark} & \multicolumn{1}{l|}{-} & \multicolumn{1}{l|}{\checkmark} & \multicolumn{1}{l|}{-} & \multicolumn{1}{l|}{-} & \multicolumn{1}{l|}{-} & \checkmark & {2} \\ 

\multicolumn{1}{l|}{} &  &  & SPIDER {\citep{yu-etal-2018-spider}} & \multicolumn{1}{l|}{-} & \multicolumn{1}{l|}{\checkmark} & \multicolumn{1}{l|}{-} & \multicolumn{1}{l|}{\checkmark} & \multicolumn{1}{l|}{\checkmark} & \multicolumn{1}{l|}{-} & \multicolumn{1}{l|}{\checkmark} & \multicolumn{1}{l|}{-} & \multicolumn{1}{l|}{-} & \multicolumn{1}{l|}{-} & \checkmark & \multicolumn{1}{l|}{-} & \multicolumn{1}{l|}{\checkmark} & \multicolumn{1}{l|}{-} & \multicolumn{1}{l|}{-} & \multicolumn{1}{l|}{\checkmark} & \multicolumn{1}{l|}{-} & \multicolumn{1}{l|}{\checkmark} & \multicolumn{1}{l|}{-} & \multicolumn{1}{l|}{-} & \multicolumn{1}{l|}{-} & \checkmark & {10} \\ 

\multicolumn{1}{l|}{} &  &  & TOP {\citep{gupta-etal-2018-semantic-parsing}} & \multicolumn{1}{l|}{-} & \multicolumn{1}{l|}{\checkmark} & \multicolumn{1}{l|}{-} & \multicolumn{1}{l|}{\checkmark} & \multicolumn{1}{l|}{\checkmark} & \multicolumn{1}{l|}{-} & \multicolumn{1}{l|}{\checkmark} & \multicolumn{1}{l|}{-} & \multicolumn{1}{l|}{\checkmark} & \multicolumn{1}{l|}{-} & - & \multicolumn{1}{l|}{-} & \multicolumn{1}{l|}{\checkmark} & \multicolumn{1}{l|}{-} & \multicolumn{1}{l|}{\checkmark} & \multicolumn{1}{l|}{-} & \multicolumn{1}{l|}{-} & \multicolumn{1}{l|}{\checkmark} & \multicolumn{1}{l|}{-} & \multicolumn{1}{l|}{\checkmark} & \multicolumn{1}{l|}{-} & - & {44} \\ \cmidrule{2-27} 

\multicolumn{1}{l|}{} & \multirow{23}{*}{\textbf{\rotatebox{90}{Response generation}}} & \multirow{4}{*}{QA} & CMUDoG {\citep{zhou-etal-2018-dataset}} & \multicolumn{1}{l|}{-} & \multicolumn{1}{l|}{\checkmark} & \multicolumn{1}{l|}{\checkmark} & \multicolumn{1}{l|}{-} & \multicolumn{1}{l|}{-} & \multicolumn{1}{l|}{\checkmark} & \multicolumn{1}{l|}{\checkmark} & \multicolumn{1}{l|}{-} & \multicolumn{1}{l|}{-} & \multicolumn{1}{l|}{\checkmark} & - & \multicolumn{1}{l|}{-} & \multicolumn{1}{l|}{\checkmark} & \multicolumn{1}{l|}{-} & \multicolumn{1}{l|}{\checkmark} & \multicolumn{1}{l|}{-} & \multicolumn{1}{l|}{-} & \multicolumn{1}{l|}{\checkmark} & \multicolumn{1}{l|}{-} & \multicolumn{1}{l|}{-} & \multicolumn{1}{l|}{\checkmark} & - & {4} \\ 

\multicolumn{1}{l|}{} &  &  & CoQA {\citep{reddy-etal-2019-coqa}} & \multicolumn{1}{l|}{-} & \multicolumn{1}{l|}{\checkmark} & \multicolumn{1}{l|}{-} & \multicolumn{1}{l|}{\checkmark} & \multicolumn{1}{l|}{-} & \multicolumn{1}{l|}{\checkmark} & \multicolumn{1}{l|}{\checkmark} & \multicolumn{1}{l|}{-} & \multicolumn{1}{l|}{-} & \multicolumn{1}{l|}{\checkmark} & - & \multicolumn{1}{l|}{-} & \multicolumn{1}{l|}{\checkmark} & \multicolumn{1}{l|}{-} & \multicolumn{1}{l|}{\checkmark} & \multicolumn{1}{l|}{-} & \multicolumn{1}{l|}{-} & \multicolumn{1}{l|}{\checkmark} & \multicolumn{1}{l|}{-} & \multicolumn{1}{l|}{-} & \multicolumn{1}{l|}{\checkmark} & - & {127} \\ 

\multicolumn{1}{l|}{} &  & & ClariQ {\citep{aliannejadi2020convai3}} & \multicolumn{1}{l|}{-} & \multicolumn{1}{l|}{\checkmark} & \multicolumn{1}{l|}{\checkmark} & \multicolumn{1}{l|}{-} & \multicolumn{1}{l|}{-} & \multicolumn{1}{l|}{\checkmark} & \multicolumn{1}{l|}{\checkmark} & \multicolumn{1}{l|}{-} & \multicolumn{1}{l|}{-} & \multicolumn{1}{l|}{\checkmark} & - & \multicolumn{1}{l|}{-} & \multicolumn{1}{l|}{\checkmark} & \multicolumn{1}{l|}{-} & \multicolumn{1}{l|}{\checkmark} & \multicolumn{1}{l|}{-} & \multicolumn{1}{l|}{-} & \multicolumn{1}{l|}{\checkmark} & \multicolumn{1}{l|}{-} & \multicolumn{1}{l|}{-} & \multicolumn{1}{l|}{\checkmark} & - & {1k} \\ 

\multicolumn{1}{l|}{} &  & & Mutual {\citep{cui-etal-2020-mutual}} & \multicolumn{1}{l|}{\checkmark} & \multicolumn{1}{l|}{-} & \multicolumn{1}{l|}{\checkmark} & \multicolumn{1}{l|}{-} & \multicolumn{1}{l|}{-} & \multicolumn{1}{l|}{\checkmark} & \multicolumn{1}{l|}{\checkmark} & \multicolumn{1}{l|}{-} & \multicolumn{1}{l|}{\checkmark} & \multicolumn{1}{l|}{-} & - & \multicolumn{1}{l|}{\checkmark} & \multicolumn{1}{l|}{-} & \multicolumn{1}{l|}{\checkmark} & \multicolumn{1}{l|}{-} & \multicolumn{1}{l|}{-} & \multicolumn{1}{l|}{-} & \multicolumn{1}{l|}{\checkmark} & \multicolumn{1}{l|}{-} & \multicolumn{1}{l|}{\checkmark} & \multicolumn{1}{l|}{\checkmark} & - & {8} \\ \cmidrule{3-27} 

\multicolumn{1}{l|}{} &  & \multirow{7}{*}{KGR}& ConvAI {\citep{yusupov-kuratov-2018-nips}} & \multicolumn{1}{l|}{-} & \multicolumn{1}{l|}{\checkmark} & \multicolumn{1}{l|}{\checkmark} & \multicolumn{1}{l|}{-} & \multicolumn{1}{l|}{-} & \multicolumn{1}{l|}{\checkmark} & \multicolumn{1}{l|}{\checkmark} & \multicolumn{1}{l|}{-} & \multicolumn{1}{l|}{-} & \multicolumn{1}{l|}{\checkmark} & - & \multicolumn{1}{l|}{-} & \multicolumn{1}{l|}{\checkmark} & \multicolumn{1}{l|}{-} & \multicolumn{1}{l|}{\checkmark} & \multicolumn{1}{l|}{-} & \multicolumn{1}{l|}{-} & \multicolumn{1}{l|}{\checkmark} & \multicolumn{1}{l|}{-} & \multicolumn{1}{l|}{-} & \multicolumn{1}{l|}{\checkmark} & - & {2} \\ 

\multicolumn{1}{l|}{} &  & & Doc2Dial {\citep{feng-etal-2020-doc2dial}} & \multicolumn{1}{l|}{-} & \multicolumn{1}{l|}{\checkmark} & \multicolumn{1}{l|}{-} & \multicolumn{1}{l|}{\checkmark} & \multicolumn{1}{l|}{-} & \multicolumn{1}{l|}{\checkmark} & \multicolumn{1}{l|}{\checkmark} & \multicolumn{1}{l|}{-} & \multicolumn{1}{l|}{-} & \multicolumn{1}{l|}{\checkmark} & - & \multicolumn{1}{l|}{-} & \multicolumn{1}{l|}{\checkmark} & \multicolumn{1}{l|}{-} & \multicolumn{1}{l|}{\checkmark} & \multicolumn{1}{l|}{-} & \multicolumn{1}{l|}{-} & \multicolumn{1}{l|}{\checkmark} & \multicolumn{1}{l|}{-} & \multicolumn{1}{l|}{-} & \multicolumn{1}{l|}{\checkmark} & - & {4} \\ 

\multicolumn{1}{l|}{} &  &  & PersonaChat {\citep{zhang-etal-2018-personalizing}} & \multicolumn{1}{l|}{\checkmark} & \multicolumn{1}{l|}{-} & \multicolumn{1}{l|}{\checkmark} & \multicolumn{1}{l|}{-} & \multicolumn{1}{l|}{-} & \multicolumn{1}{l|}{\checkmark} & \multicolumn{1}{l|}{\checkmark} & \multicolumn{1}{l|}{-} & \multicolumn{1}{l|}{-} & \multicolumn{1}{l|}{\checkmark} & - & \multicolumn{1}{l|}{\checkmark} & \multicolumn{1}{l|}{-} & \multicolumn{1}{l|}{\checkmark} & \multicolumn{1}{l|}{-} & \multicolumn{1}{l|}{-} & \multicolumn{1}{l|}{-} & \multicolumn{1}{l|}{\checkmark} & \multicolumn{1}{l|}{-} & \multicolumn{1}{l|}{-} & \multicolumn{1}{l|}{\checkmark} & - & {19} \\ 

\multicolumn{1}{l|}{} &  &  & bAbI {\citep{weston2015aicomplete}} & \multicolumn{1}{l|}{-} & \multicolumn{1}{l|}{\checkmark} & \multicolumn{1}{l|}{-} & \multicolumn{1}{l|}{\checkmark} & \multicolumn{1}{l|}{-} & \multicolumn{1}{l|}{\checkmark} & \multicolumn{1}{l|}{\checkmark} & \multicolumn{1}{l|}{-} & \multicolumn{1}{l|}{-} & \multicolumn{1}{l|}{-} & \checkmark & \multicolumn{1}{l|}{-} & \multicolumn{1}{l|}{\checkmark} & \multicolumn{1}{l|}{-} & \multicolumn{1}{l|}{\checkmark} & \multicolumn{1}{l|}{\checkmark} & \multicolumn{1}{l|}{-} & \multicolumn{1}{l|}{\checkmark} & \multicolumn{1}{l|}{-} & \multicolumn{1}{l|}{\checkmark} & \multicolumn{1}{l|}{\checkmark} & - & {161} \\ 

\multicolumn{1}{l|}{} &  &  & FaithDial {\citep{10.1162/tacl_a_00529}} & \multicolumn{1}{l|}{\checkmark} & \multicolumn{1}{l|}{-} & \multicolumn{1}{l|}{\checkmark} & \multicolumn{1}{l|}{-} & \multicolumn{1}{l|}{-} & \multicolumn{1}{l|}{\checkmark} & \multicolumn{1}{l|}{\checkmark} & \multicolumn{1}{l|}{-} & \multicolumn{1}{l|}{-} & \multicolumn{1}{l|}{\checkmark} & - & \multicolumn{1}{l|}{\checkmark} & \multicolumn{1}{l|}{-} & \multicolumn{1}{l|}{\checkmark} & \multicolumn{1}{l|}{-} & \multicolumn{1}{l|}{-} & \multicolumn{1}{l|}{\checkmark} & \multicolumn{1}{l|}{\checkmark} & \multicolumn{1}{l|}{-} & \multicolumn{1}{l|}{-} & \multicolumn{1}{l|}{\checkmark} & - & {32} \\ 

\multicolumn{1}{l|}{} &  &  & OpenDialKG {\citep{moon-etal-2019-opendialkg}} & \multicolumn{1}{l|}{-} & \multicolumn{1}{l|}{\checkmark} & \multicolumn{1}{l|}{-} & \multicolumn{1}{l|}{\checkmark} & \multicolumn{1}{l|}{-} & \multicolumn{1}{l|}{\checkmark} & \multicolumn{1}{l|}{\checkmark} & \multicolumn{1}{l|}{-} & \multicolumn{1}{l|}{-} & \multicolumn{1}{l|}{-} & \checkmark & \multicolumn{1}{l|}{-} & \multicolumn{1}{l|}{\checkmark} & \multicolumn{1}{l|}{\checkmark} & \multicolumn{1}{l|}{-} & \multicolumn{1}{l|}{-} & \multicolumn{1}{l|}{-} & \multicolumn{1}{l|}{\checkmark} & \multicolumn{1}{l|}{-} & \multicolumn{1}{l|}{-} & \multicolumn{1}{l|}{\checkmark} & - & {15} \\ 

\multicolumn{1}{l|}{} &  &  & Task2Dial {\citep{strathearn-gkatzia-2022-task2dial}} & \multicolumn{1}{l|}{-} & \multicolumn{1}{l|}{\checkmark} & \multicolumn{1}{l|}{-} & \multicolumn{1}{l|}{\checkmark} & \multicolumn{1}{l|}{-} & \multicolumn{1}{l|}{\checkmark} & \multicolumn{1}{l|}{\checkmark} & \multicolumn{1}{l|}{-} & \multicolumn{1}{l|}{-} & \multicolumn{1}{l|}{\checkmark} & - & \multicolumn{1}{l|}{-} & \multicolumn{1}{l|}{\checkmark} & \multicolumn{1}{l|}{-} & \multicolumn{1}{l|}{\checkmark} & \multicolumn{1}{l|}{\checkmark} & \multicolumn{1}{l|}{-} & \multicolumn{1}{l|}{\checkmark} & \multicolumn{1}{l|}{-} & \multicolumn{1}{l|}{-} & \multicolumn{1}{l|}{\checkmark} & - & {1} \\ \cmidrule{3-27} 

\multicolumn{1}{l|}{} &  & \multirow{6}{*}{CC} & OTTers {\citep{sevegnani-etal-2021-otters}} & \multicolumn{1}{l|}{\checkmark} & \multicolumn{1}{l|}{-} & \multicolumn{1}{l|}{\checkmark} & \multicolumn{1}{l|}{-} & \multicolumn{1}{l|}{-} & \multicolumn{1}{l|}{\checkmark} & \multicolumn{1}{l|}{\checkmark} & \multicolumn{1}{l|}{-} & \multicolumn{1}{l|}{\checkmark} & \multicolumn{1}{l|}{-} & - & \multicolumn{1}{l|}{\checkmark} & \multicolumn{1}{l|}{-} & \multicolumn{1}{l|}{\checkmark} & \multicolumn{1}{l|}{-} & \multicolumn{1}{l|}{-} & \multicolumn{1}{l|}{-} & \multicolumn{1}{l|}{\checkmark} & \multicolumn{1}{l|}{-} & \multicolumn{1}{l|}{-} & \multicolumn{1}{l|}{\checkmark} & - & {8} \\ 

\multicolumn{1}{l|}{} &  & & ProsocialDialog {\citep{kim-etal-2022-prosocialdialog}} & \multicolumn{1}{l|}{\checkmark} & \multicolumn{1}{l|}{-} & \multicolumn{1}{l|}{\checkmark} & \multicolumn{1}{l|}{-} & \multicolumn{1}{l|}{-} & \multicolumn{1}{l|}{\checkmark} & \multicolumn{1}{l|}{\checkmark} & \multicolumn{1}{l|}{-} & \multicolumn{1}{l|}{-} & \multicolumn{1}{l|}{\checkmark} & - & \multicolumn{1}{l|}{\checkmark} & \multicolumn{1}{l|}{-} & \multicolumn{1}{l|}{\checkmark} & \multicolumn{1}{l|}{\checkmark} & \multicolumn{1}{l|}{-} & \multicolumn{1}{l|}{\checkmark} & \multicolumn{1}{l|}{\checkmark} & \multicolumn{1}{l|}{-} & \multicolumn{1}{l|}{-} & \multicolumn{1}{l|}{\checkmark} & - & {5} \\ 

\multicolumn{1}{l|}{} &  &  & FusedChat {\citep{Young_Xing_Pandelea_Ni_Cambria_2022}} & \multicolumn{1}{l|}{-} & \multicolumn{1}{l|}{\checkmark} & \multicolumn{1}{l|}{-} & \multicolumn{1}{l|}{\checkmark} & \multicolumn{1}{l|}{-} & \multicolumn{1}{l|}{\checkmark} & \multicolumn{1}{l|}{\checkmark} & \multicolumn{1}{l|}{-} & \multicolumn{1}{l|}{-} & \multicolumn{1}{l|}{-} & - & \multicolumn{1}{l|}{-} & \multicolumn{1}{l|}{\checkmark} & \multicolumn{1}{l|}{-} & \multicolumn{1}{l|}{\checkmark} & \multicolumn{1}{l|}{-} & \multicolumn{1}{l|}{-} & \multicolumn{1}{l|}{\checkmark} & \multicolumn{1}{l|}{-} & \multicolumn{1}{l|}{\checkmark} & \multicolumn{1}{l|}{\checkmark} & - & {10} \\ 

\multicolumn{1}{l|}{} &  &  & mDIA {\citep{zhang2022mdia}} & \multicolumn{1}{l|}{\checkmark} & \multicolumn{1}{l|}{-} & \multicolumn{1}{l|}{\checkmark} & \multicolumn{1}{l|}{-} & \multicolumn{1}{l|}{-} & \multicolumn{1}{l|}{\checkmark} & \multicolumn{1}{l|}{\checkmark} & \multicolumn{1}{l|}{-} & \multicolumn{1}{l|}{\checkmark} & \multicolumn{1}{l|}{-} & - & \multicolumn{1}{l|}{\checkmark} & \multicolumn{1}{l|}{-} & \multicolumn{1}{l|}{\checkmark} & \multicolumn{1}{l|}{-} & \multicolumn{1}{l|}{-} & \multicolumn{1}{l|}{-} & \multicolumn{1}{l|}{\checkmark} & \multicolumn{1}{l|}{-} & \multicolumn{1}{l|}{-} & \multicolumn{1}{l|}{\checkmark} & - & {12}\\ 

\multicolumn{1}{l|}{} &  & & SODA {\citep{kim2022soda}} & \multicolumn{1}{l|}{\checkmark} & \multicolumn{1}{l|}{-} & \multicolumn{1}{l|}{\checkmark} & \multicolumn{1}{l|}{-} & \multicolumn{1}{l|}{-} & \multicolumn{1}{l|}{\checkmark} & \multicolumn{1}{l|}{\checkmark} & \multicolumn{1}{l|}{-} & \multicolumn{1}{l|}{-} & \multicolumn{1}{l|}{\checkmark} & - & \multicolumn{1}{l|}{\checkmark} & \multicolumn{1}{l|}{-} & \multicolumn{1}{l|}{\checkmark} & \multicolumn{1}{l|}{\checkmark} & \multicolumn{1}{l|}{-} & \multicolumn{1}{l|}{\checkmark} & \multicolumn{1}{l|}{\checkmark} & \multicolumn{1}{l|}{-} & \multicolumn{1}{l|}{\checkmark} & \multicolumn{1}{l|}{\checkmark} & - & {1k} \\ 

\multicolumn{1}{l|}{} &  &  & Switchboard-1 {\citep{Jurafsky-etal:1997}} & \multicolumn{1}{l|}{\checkmark} & \multicolumn{1}{l|}{-} & \multicolumn{1}{l|}{\checkmark} & \multicolumn{1}{l|}{-} & \multicolumn{1}{l|}{-} & \multicolumn{1}{l|}{\checkmark} & \multicolumn{1}{l|}{\checkmark} & \multicolumn{1}{l|}{-} & \multicolumn{1}{l|}{\checkmark} & \multicolumn{1}{l|}{-} & - & \multicolumn{1}{l|}{\checkmark} & \multicolumn{1}{l|}{-} & \multicolumn{1}{l|}{\checkmark} & \multicolumn{1}{l|}{-} & \multicolumn{1}{l|}{-} & \multicolumn{1}{l|}{-} & \multicolumn{1}{l|}{\checkmark} & \multicolumn{1}{l|}{-} & \multicolumn{1}{l|}{\checkmark} & \multicolumn{1}{l|}{\checkmark} & - & {2} \\ \cmidrule{3-27} 

\multicolumn{1}{l|}{} &  & \multirow{6}{*}{TOD} & Ubuntu Dialogue Corpus {\citep{lowe-etal-2015-ubuntu}} & \multicolumn{1}{l|}{-} & \multicolumn{1}{l|}{\checkmark} & \multicolumn{1}{l|}{-} & \multicolumn{1}{l|}{\checkmark} & \multicolumn{1}{l|}{-} & \multicolumn{1}{l|}{\checkmark} & \multicolumn{1}{l|}{\checkmark} & \multicolumn{1}{l|}{-} & \multicolumn{1}{l|}{\checkmark} & \multicolumn{1}{l|}{-} & - & \multicolumn{1}{l|}{-} & \multicolumn{1}{l|}{\checkmark} & \multicolumn{1}{l|}{-} & \multicolumn{1}{l|}{\checkmark} & \multicolumn{1}{l|}{\checkmark} & \multicolumn{1}{l|}{-} & \multicolumn{1}{l|}{\checkmark} & \multicolumn{1}{l|}{-} & \multicolumn{1}{l|}{-} & \multicolumn{1}{l|}{\checkmark} & - & {1k} \\ 

\multicolumn{1}{l|}{} &  &  & ABCD {\citep{chen-etal-2021-action}} & \multicolumn{1}{l|}{-} & \multicolumn{1}{l|}{\checkmark} & \multicolumn{1}{l|}{-} & \multicolumn{1}{l|}{\checkmark} & \multicolumn{1}{l|}{-} & \multicolumn{1}{l|}{\checkmark} & \multicolumn{1}{l|}{\checkmark} & \multicolumn{1}{l|}{-} & \multicolumn{1}{l|}{-} & \multicolumn{1}{l|}{-} & \checkmark & \multicolumn{1}{l|}{-} & \multicolumn{1}{l|}{\checkmark} & \multicolumn{1}{l|}{-} & \multicolumn{1}{l|}{\checkmark} & \multicolumn{1}{l|}{\checkmark} & \multicolumn{1}{l|}{-} & \multicolumn{1}{l|}{\checkmark} & \multicolumn{1}{l|}{-} & \multicolumn{1}{l|}{-} & \multicolumn{1}{l|}{\checkmark} & - & {10} \\ 

\multicolumn{1}{l|}{} &  & & BiTOD {\citep{lin2021bitod}} & \multicolumn{1}{l|}{-} & \multicolumn{1}{l|}{\checkmark} & \multicolumn{1}{l|}{-} & \multicolumn{1}{l|}{\checkmark} & \multicolumn{1}{l|}{-} & \multicolumn{1}{l|}{\checkmark} & \multicolumn{1}{l|}{\checkmark} & \multicolumn{1}{l|}{-} & \multicolumn{1}{l|}{\checkmark} & \multicolumn{1}{l|}{-} & - & \multicolumn{1}{l|}{-} & \multicolumn{1}{l|}{\checkmark} & \multicolumn{1}{l|}{-} & \multicolumn{1}{l|}{\checkmark} & \multicolumn{1}{l|}{-} & \multicolumn{1}{l|}{-} & \multicolumn{1}{l|}{\checkmark} & \multicolumn{1}{l|}{-} & \multicolumn{1}{l|}{\checkmark} & \multicolumn{1}{l|}{\checkmark} & - & {7} \\ 

\multicolumn{1}{l|}{} &  & & CraiglistBargains {\citep{he-etal-2018-decoupling}} & \multicolumn{1}{l|}{-} & \multicolumn{1}{l|}{\checkmark} & \multicolumn{1}{l|}{-} & \multicolumn{1}{l|}{\checkmark} & \multicolumn{1}{l|}{-} & \multicolumn{1}{l|}{\checkmark} & \multicolumn{1}{l|}{\checkmark} & \multicolumn{1}{l|}{-} & \multicolumn{1}{l|}{-} & \multicolumn{1}{l|}{-} & \checkmark & \multicolumn{1}{l|}{-} & \multicolumn{1}{l|}{\checkmark} & \multicolumn{1}{l|}{\checkmark} & \multicolumn{1}{l|}{\checkmark} & \multicolumn{1}{l|}{-} & \multicolumn{1}{l|}{-} & \multicolumn{1}{l|}{\checkmark} & \multicolumn{1}{l|}{-} & \multicolumn{1}{l|}{-} & \multicolumn{1}{l|}{\checkmark} & - & {6} \\ 

\multicolumn{1}{l|}{} &  &  & DeliData {\citep{Karadzhov2021DeliDataAD}} & \multicolumn{1}{l|}{-} & \multicolumn{1}{l|}{\checkmark} & \multicolumn{1}{l|}{\checkmark} & \multicolumn{1}{l|}{-} & \multicolumn{1}{l|}{-} & \multicolumn{1}{l|}{\checkmark} & \multicolumn{1}{l|}{\checkmark} & \multicolumn{1}{l|}{-} & \multicolumn{1}{l|}{-} & \multicolumn{1}{l|}{-} & \checkmark & \multicolumn{1}{l|}{-} & \multicolumn{1}{l|}{\checkmark} & \multicolumn{1}{l|}{-} & \multicolumn{1}{l|}{\checkmark} & \multicolumn{1}{l|}{\checkmark} & \multicolumn{1}{l|}{-} & \multicolumn{1}{l|}{\checkmark} & \multicolumn{1}{l|}{-} & \multicolumn{1}{l|}{-} & \multicolumn{1}{l|}{\checkmark} & - & {0.5} \\ 

\multicolumn{1}{l|}{} &  &  & MetalWOz {\citep{shalyminov-etal-2019-shot}} & \multicolumn{1}{l|}{-} & \multicolumn{1}{l|}{\checkmark} & \multicolumn{1}{l|}{-} & \multicolumn{1}{l|}{\checkmark} & \multicolumn{1}{l|}{-} & \multicolumn{1}{l|}{\checkmark} & \multicolumn{1}{l|}{\checkmark} & \multicolumn{1}{l|}{-} & \multicolumn{1}{l|}{\checkmark} & \multicolumn{1}{l|}{-} & - & \multicolumn{1}{l|}{-} & \multicolumn{1}{l|}{\checkmark} & \multicolumn{1}{l|}{-} & \multicolumn{1}{l|}{\checkmark} & \multicolumn{1}{l|}{-} & \multicolumn{1}{l|}{-} & \multicolumn{1}{l|}{\checkmark} & \multicolumn{1}{l|}{-} & \multicolumn{1}{l|}{-} & \multicolumn{1}{l|}{\checkmark} & - & {10} \\ \midrule

\multicolumn{2}{l|}{\multirow{10}{*}{\textbf{\rotatebox{90}{Classification}}}} & \multirow{4}{*}{ID} & Banking77 {\citep{casanueva-etal-2020-efficient}} & \multicolumn{1}{l|}{-} & \multicolumn{1}{l|}{\checkmark} & \multicolumn{1}{l|}{-} & \multicolumn{1}{l|}{\checkmark} & \multicolumn{1}{l|}{\checkmark} & \multicolumn{1}{l|}{-} & \multicolumn{1}{l|}{\checkmark} & \multicolumn{1}{l|}{-} & \multicolumn{1}{l|}{\checkmark} & \multicolumn{1}{l|}{-} & - & \multicolumn{1}{l|}{-} & \multicolumn{1}{l|}{\checkmark} & \multicolumn{1}{l|}{-} & \multicolumn{1}{l|}{\checkmark} & \multicolumn{1}{l|}{-} & \multicolumn{1}{l|}{-} & \multicolumn{1}{l|}{\checkmark} & \multicolumn{1}{l|}{-} & \multicolumn{1}{l|}{\checkmark} & \multicolumn{1}{l|}{-} & - & {13} \\ 

\multicolumn{2}{l|}{} & & CLINC150 {\citep{larson-etal-2019-evaluation}} & \multicolumn{1}{l|}{-} & \multicolumn{1}{l|}{\checkmark} & \multicolumn{1}{l|}{-} & \multicolumn{1}{l|}{\checkmark} & \multicolumn{1}{l|}{\checkmark} & \multicolumn{1}{l|}{-} & \multicolumn{1}{l|}{\checkmark} & \multicolumn{1}{l|}{-} & \multicolumn{1}{l|}{\checkmark} & \multicolumn{1}{l|}{-} & - & \multicolumn{1}{l|}{-} & \multicolumn{1}{l|}{\checkmark} & \multicolumn{1}{l|}{-} & \multicolumn{1}{l|}{\checkmark} & \multicolumn{1}{l|}{-} & \multicolumn{1}{l|}{-} & \multicolumn{1}{l|}{\checkmark} & \multicolumn{1}{l|}{-} & \multicolumn{1}{l|}{\checkmark} & \multicolumn{1}{l|}{-} & - & {23} \\ 

\multicolumn{2}{l|}{} & & HWU64 {\citep{Liu2021}} & \multicolumn{1}{l|}{-} & \multicolumn{1}{l|}{\checkmark} & \multicolumn{1}{l|}{-} & \multicolumn{1}{l|}{\checkmark} & \multicolumn{1}{l|}{\checkmark} & \multicolumn{1}{l|}{-} & \multicolumn{1}{l|}{\checkmark} & \multicolumn{1}{l|}{-} & \multicolumn{1}{l|}{\checkmark} & \multicolumn{1}{l|}{-} & - & \multicolumn{1}{l|}{-} & \multicolumn{1}{l|}{\checkmark} & \multicolumn{1}{l|}{-} & \multicolumn{1}{l|}{\checkmark} & \multicolumn{1}{l|}{-} & \multicolumn{1}{l|}{-} & \multicolumn{1}{l|}{\checkmark} & \multicolumn{1}{l|}{-} & \multicolumn{1}{l|}{\checkmark} & \multicolumn{1}{l|}{-} & - & {11} \\ 

\multicolumn{2}{l|}{} &  & SGD {\citep{Rastogi_Zang_Sunkara_Gupta_Khaitan_2020}} & \multicolumn{1}{l|}{-} & \multicolumn{1}{l|}{\checkmark} & \multicolumn{1}{l|}{-} & \multicolumn{1}{l|}{\checkmark} & \multicolumn{1}{l|}{\checkmark} & \multicolumn{1}{l|}{-} & \multicolumn{1}{l|}{\checkmark} & \multicolumn{1}{l|}{-} & \multicolumn{1}{l|}{\checkmark} & \multicolumn{1}{l|}{-} & - & \multicolumn{1}{l|}{-} & \multicolumn{1}{l|}{\checkmark} & \multicolumn{1}{l|}{-} & \multicolumn{1}{l|}{\checkmark} & \multicolumn{1}{l|}{-} & \multicolumn{1}{l|}{-} & \multicolumn{1}{l|}{\checkmark} & \multicolumn{1}{l|}{-} & \multicolumn{1}{l|}{\checkmark} & \multicolumn{1}{l|}{-} & - & {16} \\ \cmidrule{3-27} 

\multicolumn{2}{l|}{} & SF & Restaurant8k {\citep{coope-etal-2020-span}} & \multicolumn{1}{l|}{-} & \multicolumn{1}{l|}{\checkmark} & \multicolumn{1}{l|}{-} & \multicolumn{1}{l|}{\checkmark} & \multicolumn{1}{l|}{\checkmark} & \multicolumn{1}{l|}{-} & \multicolumn{1}{l|}{\checkmark} & \multicolumn{1}{l|}{-} & \multicolumn{1}{l|}{\checkmark} & \multicolumn{1}{l|}{-} & - & \multicolumn{1}{l|}{-} & \multicolumn{1}{l|}{\checkmark} & \multicolumn{1}{l|}{-} & \multicolumn{1}{l|}{\checkmark} & \multicolumn{1}{l|}{-} & \multicolumn{1}{l|}{-} & \multicolumn{1}{l|}{\checkmark} & \multicolumn{1}{l|}{-} & \multicolumn{1}{l|}{\checkmark} & \multicolumn{1}{l|}{-} & - & {11} \\ \cmidrule{3-27} 

\multicolumn{2}{l|}{} & DST & MultiWOZ2.1 {\citep{eric-etal-2020-multiwoz}} & \multicolumn{1}{l|}{-} & \multicolumn{1}{l|}{\checkmark} & \multicolumn{1}{l|}{-} & \multicolumn{1}{l|}{\checkmark} & \multicolumn{1}{l|}{-} & \multicolumn{1}{l|}{\checkmark} & \multicolumn{1}{l|}{\checkmark} & \multicolumn{1}{l|}{-} & \multicolumn{1}{l|}{\checkmark} & \multicolumn{1}{l|}{-} & - & \multicolumn{1}{l|}{-} & \multicolumn{1}{l|}{\checkmark} & \multicolumn{1}{l|}{-} & \multicolumn{1}{l|}{\checkmark} & \multicolumn{1}{l|}{\checkmark} & \multicolumn{1}{l|}{-} & \multicolumn{1}{l|}{\checkmark} & \multicolumn{1}{l|}{-} & \multicolumn{1}{l|}{\checkmark} & \multicolumn{1}{l|}{\checkmark} & - & {10} \\ \cmidrule{3-27} 

\multicolumn{2}{l|}{} & \multirow{4}{*}{AD} & DailyDialogue {\citep{li-etal-2017-dailydialog}} & \multicolumn{1}{l|}{\checkmark} & \multicolumn{1}{l|}{-} & \multicolumn{1}{l|}{\checkmark} & \multicolumn{1}{l|}{-} & \multicolumn{1}{l|}{-} & \multicolumn{1}{l|}{\checkmark} & \multicolumn{1}{l|}{\checkmark} & \multicolumn{1}{l|}{-} & \multicolumn{1}{l|}{\checkmark} & \multicolumn{1}{l|}{-} & - & \multicolumn{1}{l|}{\checkmark} & \multicolumn{1}{l|}{-} & \multicolumn{1}{l|}{\checkmark} & \multicolumn{1}{l|}{-} & \multicolumn{1}{l|}{-} & \multicolumn{1}{l|}{\checkmark} & \multicolumn{1}{l|}{\checkmark} & \multicolumn{1}{l|}{-} & \multicolumn{1}{l|}{\checkmark} & \multicolumn{1}{l|}{\checkmark} & - & {11} \\ 

\multicolumn{2}{l|}{} &  & MELD {\citep{poria-etal-2019-meld}} & \multicolumn{1}{l|}{\checkmark} & \multicolumn{1}{l|}{-} & \multicolumn{1}{l|}{\checkmark} & \multicolumn{1}{l|}{-} & \multicolumn{1}{l|}{-} & \multicolumn{1}{l|}{\checkmark} & \multicolumn{1}{l|}{-} & \multicolumn{1}{l|}{\checkmark} & \multicolumn{1}{l|}{\checkmark} & \multicolumn{1}{l|}{-} & - & \multicolumn{1}{l|}{\checkmark} & \multicolumn{1}{l|}{-} & \multicolumn{1}{l|}{\checkmark} & \multicolumn{1}{l|}{-} & \multicolumn{1}{l|}{-} & \multicolumn{1}{l|}{\checkmark} & \multicolumn{1}{l|}{\checkmark} & \multicolumn{1}{l|}{-} & \multicolumn{1}{l|}{\checkmark} & \multicolumn{1}{l|}{\checkmark} & - & {1} \\ 

\multicolumn{2}{l|}{} &  & MUStARD {\citep{castro-etal-2019-towards}} & \multicolumn{1}{l|}{\checkmark} & \multicolumn{1}{l|}{-} & \multicolumn{1}{l|}{\checkmark} & \multicolumn{1}{l|}{-} & \multicolumn{1}{l|}{-} & \multicolumn{1}{l|}{\checkmark} & \multicolumn{1}{l|}{-} & \multicolumn{1}{l|}{\checkmark} & \multicolumn{1}{l|}{\checkmark} & \multicolumn{1}{l|}{-} & - & \multicolumn{1}{l|}{\checkmark} & \multicolumn{1}{l|}{-} & \multicolumn{1}{l|}{\checkmark} & \multicolumn{1}{l|}{-} & \multicolumn{1}{l|}{-} & \multicolumn{1}{l|}{-} & \multicolumn{1}{l|}{\checkmark} & \multicolumn{1}{l|}{-} & \multicolumn{1}{l|}{\checkmark} & \multicolumn{1}{l|}{\checkmark} & - & {6} \\ 

\multicolumn{2}{l|}{} &  & Empathetic Dialogues {\citep{Rashkin2018TowardsEO}} & \multicolumn{1}{l|}{\checkmark} & \multicolumn{1}{l|}{-} & \multicolumn{1}{l|}{\checkmark} & \multicolumn{1}{l|}{-} & \multicolumn{1}{l|}{-} & \multicolumn{1}{l|}{\checkmark} & \multicolumn{1}{l|}{\checkmark} & \multicolumn{1}{l|}{-} & \multicolumn{1}{l|}{-} & \multicolumn{1}{l|}{-} & - & \multicolumn{1}{l|}{\checkmark} & \multicolumn{1}{l|}{-} & \multicolumn{1}{l|}{\checkmark} & \multicolumn{1}{l|}{-} & \multicolumn{1}{l|}{-} & \multicolumn{1}{l|}{\checkmark} & \multicolumn{1}{l|}{\checkmark} & \multicolumn{1}{l|}{-} & \multicolumn{1}{l|}{-} & \multicolumn{1}{l|}{\checkmark} & - & {24} \\ \bottomrule
\end{tabular}%
}
\caption{Characteristic of each task based on the taxonomic characteristic of a dialogue agent. {Size indicates an approximate value expressed in thousands (k).} Abbreviations -- DR: Dialogue Rewrite, DS: Dialogue Summary, D2S: Dialogue to Structure, QA: Question Answering, KGR: Knowledge Grounded Response, CC: Chit-chat, TOD: Task Oriented Dialogues, ID: Intent Detection, SF: Slot Filling, DST: Dialogue State Tracking, AD: Affect Detection, CC: Chit-chat, GO: Goal Oriented, Spc: Specific, ST: Single Turn, MT: Multi Turn, U: Unimodal, M: Multimodal, Unstr: Unstructured, Str: Structured, Eng: Engaging, Inf: Informative, Instr: Instructional, Emp: Empathetic.}
\label{tab:tasks_taxonomy}
\vspace{-4mm}
\end{table*}

By drawing upon the taxonomy depicted in Figure \ref{fig:taxonomy} and existing literature, we identify \textit{eleven} distinct tasks related to dialogue that capture all necessary characteristics of a dialogue agent. In order to construct a dialogue agent, a practitioner must be aware of these tasks, which can be classified into two primary categories -- generative and classification. Specifically, the identified tasks include \textbf{Dialogue Rewrite (DR)}  \citep{elgohary-etal-2019-unpack},\textbf{ Dialogue Summary (DS)} \citep{gliwa-etal-2019-samsum,chen-etal-2021-dialogsum}, \textbf{Dialogue to Structure (D2S)} \citep{yu-etal-2019-cosql,yu-etal-2018-spider,gupta-etal-2018-semantic-parsing}, \textbf{Question Answering(QA)} \citep{zhou-etal-2018-dataset,reddy-etal-2019-coqa,aliannejadi2020convai3,cui-etal-2020-mutual}, \textbf{Knowledge Grounded Response (KGR)} \citep{yusupov-kuratov-2018-nips,feng-etal-2020-doc2dial,zhang-etal-2018-personalizing,weston2015aicomplete,10.1162/tacl_a_00529,moon-etal-2019-opendialkg,strathearn-gkatzia-2022-task2dial}, \textbf{Chit-Chat (CC)} \citep{sevegnani-etal-2021-otters,kim-etal-2022-prosocialdialog,Young_Xing_Pandelea_Ni_Cambria_2022,zhang2022mdia,kim2022soda,Jurafsky-etal:1997}, and \textbf{Task-Oriented Dialogues (TOD)} \citep{lowe-etal-2015-ubuntu,chen-etal-2021-action,weston2015aicomplete,lin2021bitod,he-etal-2018-decoupling,Karadzhov2021DeliDataAD,shalyminov-etal-2019-shot} in the generative category, and \textbf{Intent Detection(ID)} \citep{casanueva-etal-2020-efficient,larson-etal-2019-evaluation,Liu2021,Rastogi_Zang_Sunkara_Gupta_Khaitan_2020}, \textbf{Slot Filling (SF)} \citep{coope-etal-2020-span}, \textbf{Dialogue State Tracking (DST)} \citep{eric-etal-2020-multiwoz}, and \textbf{Affect Detection (AD)} \citep{poria-etal-2019-meld,li-etal-2017-dailydialog,castro-etal-2019-towards,rashkin-etal-2019-towards} in the classification category. Table \ref{tab:tasks_taxonomy} summarises all the datasets considered in this study for each of the mentioned tasks and illustrates the characteristics satisfied by each of these tasks from the taxonomy.
{As we delve into the details of each task type in the forthcoming sections, it is noteworthy to highlight a few observations obtained from the presented table.
\begin{itemize}[noitemsep,topsep=0pt]
    \item In dialogue datasets featuring chit-chat conversations, an inclination towards characteristics indicative of open domain, multi-turn interactions, and the absence of external knowledge is observed. Notably, a prevalent trend emerges in the generation of similar output within such datasets. An identified gap in the existing landscape pertains to the scarcity of datasets integrating external knowledge with chit-chat dialogues. Recognizing the potential enrichment that associated knowledge, particularly commonsense \citep{ghosal-etal-2020-cosmic}, can bring to dialogues, it becomes a potential future research area.
    \item For instances where the dataset comprises goal-oriented conversations, it is probable that the dataset is tailored to a specific domain, assisted with either structured or unstructured knowledge linked to it. Goal-oriented dialogues typically centre around specific tasks like booking airline tickets, scheduling doctor appointments, or securing restaurant reservations. Notably, these `goals' can extend beyond specific tasks to encompass aspects such as the accomplishment of the goal of dialogue engagement \citep{gottardi2022alexa}. Intriguingly, such goal-orientation does not necessarily confine the dialogue to a predefined domain, allowing for an open-domain context. A prospective avenue for research lies in the development of more open-domain, goal-oriented dialogue datasets that focus more on conversational goals like user engagement.
    \item The chit-chat setting exhibits the predominant trend of producing extensive and engaging dialogue output \citep{gottardi2022alexa}. In contrast, the goal-oriented setting commonly yields responses characterized by informativeness, instructional clarity, and brevity \citep{muise2019planning}. Intriguingly, datasets combining both goal-oriented and chit-chat conversations are notably sparse, despite real-world dialogues frequently encompassing a fluid interchange between these conversational types \citep{shuster2022blenderbot}. The presence of such datasets could substantially enhance the research community's capabilities and insights.
\end{itemize}
}

\subsection{Generative Dialogue Tasks}
Generative dialogue tasks require the handling of diverse input and output characteristics \citep{chen2017survey}. These tasks can be classified into two distinct types -- transformation and response generation. In transformation tasks, the output of the given input conversation is not the subsequent response but rather some other meaningful text, such as a dialogue summary \citep{gliwa-etal-2019-samsum}. On the other hand, response generation tasks involve generating the next response in the dialogue, given an input context \citep{zhang-etal-2020-dialogpt}.

\subsubsection{Transformation Tasks}
\paragraph{Dialogue Rewrite (DR)} {This task involves} the challenging process of modifying a given conversational utterance to better fit a specific social context or conversational objective, while retaining its original meaning. To explore this task further, we turn to the CANARD dataset \citep{elgohary-etal-2019-unpack}.
This dataset is specifically designed for rewriting context-dependent questions into self-contained questions that can be answered independently by resolving all coreferences.
The objective is to ensure that the new question has the same answer as the original one. 
\citet{quan-etal-2019-gecor} and \citet{martin-etal-2020-mudoco} proposed the TASK and MuDoCo datasets, respectively, focusing on rewriting dialogues in a way that coreferences and ellipsis are resolved.
\citet{Huang_Li_Zou_Zhang_2021} combined sequence labelling and autoregression techniques to restore utterances without any coreferences. In contrast, \citet{article_Durese} shaped the dialogue rewrite task as sentence editing and predicted edit operations for each word in the context. Other methods also use knowledge augmentation \citep{ke-etal-2022-knowledge}, reinforcement learning \citep{chen2022reinforced}, and the copy mechanism \citep{quan-etal-2019-gecor}.

\begin{mdframed}
        \underline{\textit{Key challenges.}} Despite achieving a reasonable performance in the dialogue rewrite task, some challenges remain, with the major obstacle being the inclusion of new words in the ground truth annotations that are difficult to incorporate into the predicted rewrite \citep{liu-etal-2020-incomplete}. { In order to mitigate this challenge, many studies have explored the methods of lexicon integration \citep{lee2023orchestrallm, czarnowska-etal-2020-morphologically}, open-vocabulary \citep{10.5555/3455716.3455856, hao-etal-2021-rast, vu-etal-2022-overcoming}, and context-aware encoding \citep{xiao2020convolutional, vinyals2015order}.}
\end{mdframed}

\paragraph{Dialogue summary (DS)}
Dialogues, despite their importance in communication, can often become lengthy and veer off-topic. This can make it challenging to extract the meaningful content from the entire conversation. To overcome this issue, the task of dialogue summarization has emerged. 
Dialogue summarization presents a concise account of the key topics, ideas, and arguments discussed during the conversation. There are two prominent datasets that address the challenge of dialogue summarization: the SAMSum \citep{gliwa-etal-2019-samsum} and DialogSum \citep{chen-etal-2021-dialogsum} corpora consisting of dialogues and their corresponding summaries.
The SAMSum dataset consists of dialogues that were curated by linguists who are fluent in English and who attempted to simulate messenger-like conversations. While DialogSum consists of face-to-face spoken dialogues covering various daily life topics such as schooling, work, and shopping. The dialogues are present in the textual format in both datasets.
{Other datasets such as QMSum \citep{zhong-etal-2021-qmsum}, MediaSum \citep{zhu-etal-2021-mediasum}, DiDi \citep{10.1145/3292500.3330683}, CCCS \citep{favre-etal-2015-call}, Telemedicine \citep{joshi-etal-2020-dr}, CRD3 \citep{rameshkumar-bailey-2020-storytelling}, Television Shows \citep{zechner-waibel-2000-diasumm}, AutoMin \citep{nedoluzhko-etal-2022-elitr}, and Clinical Encounter Visits \citep{yim-yetisgen-2021-towards} are also constructed for the task of dialogue summarisation. For a detailed guide on the task, we redirect the readers to the extensive survey conducted by \citet{tuggener-etal-2021-summarizing}. Many architectures have been proposed to solve the task of dialogue summarisation.}
\citet{liang2023enhancing} uses topic-aware Global-Local Centrality (GLC) to extract important context from all sub-topics.
By combining global- and local-level centralities, the GLC method guides the model to capture salient context and sub-topics while generating summaries. 
Other studies have utilized contrastive loss \citep{10055286}, multi-view summary generation \citep{Chen2020MultiViewSM}, post-processing techniques improving the quality of summaries \citep{Lee2021CapturingSI}, external knowledge incorporation \citep{kim-etal-2022-mind}, multimodal summarisation \citep{10.1016/j.knosys.2021.107152}, and methods to reduce hallucinations in generated summaries \citep{Liu2021ControllableND, Narayan2021PlanningWL, Wu2021ControllableAD}.

\begin{mdframed}
        \underline{\textit{Key challenges.}} With the help of pre-trained language models, current methods are adept at converting the original chat into a concise summary. Nonetheless, these models still face challenges in selecting the crucial parts and tend to generate hallucinations \citep{feng2022survey}. In the case of longer dialogues, the models may exhibit bias towards a specific part of the chat, such as the beginning or end, producing summaries that are not entirely satisfactory \citep{dey-etal-2020-corpora}. { Many studies explore novel attention mechanism with topic modeling \citep{xiao2020convolutional}, reinforcement learning and differential rewards \citep{chen-etal-2023-human, ITALIANI2024127132, zhang-etal-2023-macsum}, and knowledge augmentation with fact-checking \citep{hua-etal-2023-improving, hwang-etal-2023-dialogizer} to mitigate these challenges.}
\end{mdframed}

\paragraph{Dialogue to structure (D2S)}
Although natural language is the fundamental way humans communicate, the interaction between humans and machines often requires a more structured language such as SQL or syntactic trees. Tasks such as \textit{Text-to-SQL} and \textit{Semantic Parsing} seek to bridge the gap between natural language and machine-understandable forms of communication.
To address this, four prominent datasets have been developed -- CoSQL \citep{yu-etal-2019-cosql}, SPIDER \citep{yu-etal-2018-spider}, and WikiSQL \citep{zhongSeq2SQL2017} for text-to-sql, which are composed of pairs of natural language queries paired with their corresponding SQL queries, and the Task Oriented Parsing (TOP) dataset \citep{gupta-etal-2018-semantic-parsing} for semantic parsing which contains conversations that are annotated with hierarchical semantic representation for task-oriented dialogue systems. There are numerous approaches to handling these datasets, including encoder/decoder models with decoder constraints \citep{Wang2019RATSQLRS, yin-neubig-2017-syntactic}, large language models without any constraints \citep{Suhr2020ExploringUG, Lin2020BridgingTA}, final hypothesis pruning \citep{Scholak2021PICARDPI}, span-based extraction \citep{Pasupat2019SpanbasedHS, Meng2022LexiconinjectedSP}, data augmentation \citep{Lee2022AugmentingTD, Xuan2020ImprovingSS}, and ensembling techniques \citep{Einolghozati2019ImprovingSP}.

\begin{mdframed}
       \underline{\textit{Key challenges.}} Despite recent advancements in D2S type tasks, there remains a scarcity of high-quality resources related to complex queries \citep{Lee2022AugmentingTD}. Furthermore, the performance of D2S models tends to be suboptimal when encountering small perturbations, such as synonym substitutions or the introduction of domain-specific knowledge in the input \citep{qin2022survey}. { Existing studies explore the areas of data augmentation with resource creation to solve this challenge \citep{joshi-etal-2022-augmenting, visualbias-inproceedings}. Enhancing robustness and handling perturbation \citep{yu2023chain, 10.1145/3292500.3330957} are other possible solutions to the challenge of brittleness in the D2S tasks.} Further research in this direction could yield valuable insights. 
\end{mdframed}

\subsubsection{Response Generation}
\paragraph{Question Answering (QA)}
Dialogue agents must possess the ability to ask relevant questions {in order to engage the participants by introducing interesting topics via questions in general chit-chat setting \citep{gottardi2022alexa},} and provide appropriate answers to user inquiries{, to remain authentic in the QA setting \citep{elgohary-etal-2019-unpack}}.
As a result, Question Answering (QA) is a crucial task for dialogue agents to perform competently. To this end, datasets such as CMUDoG \citep{zhou-etal-2018-dataset}, CoQA \citep{reddy-etal-2019-coqa}, SQuAD \citep{rajpurkar-etal-2016-squad, rajpurkar-etal-2018-know}, ClariQ \citep{aliannejadi2020convai3}, and Mutual \citep{cui-etal-2020-mutual} are among the most notable and widely used for the purpose of training and evaluating QA systems. If external knowledge is used to answer questions, the task can be termed as knowledge-grounded question answering \citep{10.1145/3397271.3401097}. The CMUDoG, CoQA, and SQuAD datasets are examples of this category. The FIRE model \citep{gu-etal-2020-filtering} utilizes context and knowledge filters to create context- and knowledge-aware representations through global and bidirectional attention. Other methods include multitask learning \citep{zhou2020multidomain}, semantic parsing \citep{berant-liang-2014-semantic,reddy-etal-2014-large}, knowledge-based grounding \citep{yih-etal-2015-semantic,liang-etal-2017-neural}, and information-retrieval based methods \citep{bordes2015largescale,dong-etal-2015-question}. On the other hand, the ClariQ and Mutual datasets does not contain any external knowledge. \citet{komeili-etal-2022-internet} have proposed using the Internet as a source for obtaining relevant information. In contrast, \citet{hixon-etal-2015-learning} proposes to learn domain from conversation context. Zero-shot approaches \citep{wang2023zeroshot}, adversarial pretraining \citep{pi2022logigan}, convolution networks \citep{liu2022dialogconv}, and graph based methods \citep{ouyang-etal-2021-dialogue} are also used to solve the task of QA.

\begin{mdframed}
       \underline{\textit{Key challenges.}} In the field of discourse-based question answering, which requires models to consider both deep conversation context and potential external knowledge, anaphora resolution still poses a significant challenge that necessitates further investigation \citep{pandya2021question}. Additionally, capturing long dialogue context \citep{christmann2022conversational} and preventing topical drift \citep{venkataram2020topiqal} offers other research direction. { Many studies explore these challenges and propose viable solutions to mitigate them \citep{lin-etal-2021-contextualized, wu-etal-2023-multi-task}. However, a reliable solution still needs more research in the field.}
\end{mdframed}

\paragraph{Knowledge grounded response (KGR)}
Similar to knowledge-grounded question answering, knowledge-grounded response generation is a task that utilizes external knowledge to generate relevant responses. Some of the primary datasets related to knowledge grounding include ConvAI \citep{yusupov-kuratov-2018-nips}, Doc2Dial \citep{feng-etal-2020-doc2dial}, PersonaChat \citep{zhang-etal-2018-personalizing}, bAbI \citep{weston2015aicomplete}, FaithDial \citep{10.1162/tacl_a_00529}, OpenDialKG \citep{moon-etal-2019-opendialkg}, and Task2Dial \citep{strathearn-gkatzia-2022-task2dial}. Most methods that aim to solve the task of knowledge grounded response generation, like knowledge grounded QA, uses a two step approach of retrieval and generation \citep{zhan-etal-2021-augmenting, Wu_Galley_Brockett_2021}, graph-based approach \citep{9207054, 10.1145/3404835.3463000},  reinforcement learning approach \citep{hedayatnia-etal-2020-policy}, and {retrieval}-free approaches \citep{xu-etal-2022-retrieval}.

\begin{mdframed}
       \underline{\textit{Key challenges.}} The current trend in knowledge grounded response generation is to use a two-step approach of retrieval and generation, which increases the complexity of the system \citep{zhou2022think}. Recently, researchers such as \citet{xu-etal-2022-retrieval} and \citet{zhou2022think} have explored ways to bypass the retrieval step and produce more efficient models. Further research in this direction can improve the efficiency of systems.
\end{mdframed}

\paragraph{Chit-chat (CC)}
The primary goal of a dialogue agent is to generate responses, whether it is for chit-chat based dialogues or task-oriented dialogues. This section will specifically focus on the response generation for chit-chat agents. While there are numerous dialogue datasets available that contain chit-chat dialogues and can be used as training data, such as PersonaChat {\citep{zhang-etal-2018-personalizing}}, MELD {\citep{poria-etal-2019-meld}}, DailyDialogue {\citep{li-etal-2017-dailydialog}}, MUStARD {\citep{castro-etal-2019-towards}}, and Mutual {\citep{cui-etal-2020-mutual}}, there are some datasets specifically curated for the task of chit-chat generation. Examples of such datasets include OTTers \citep{sevegnani-etal-2021-otters}, ProsocialDialog \citep{kim-etal-2022-prosocialdialog}, FusedChat \citep{Young_Xing_Pandelea_Ni_Cambria_2022}, mDIA \citep{zhang2022mdia}, SODA \citep{kim2022soda}, and the Switchboard-1 corpus \citep{Jurafsky-etal:1997}. Major approaches used to generate responses for chit-chat dialogue agents include the use of contrastive learning \citep{Li2022MitigatingNS, li2021enhancing, cai2020group}, continual learning \citep{Liu2022WhereTG,liu2021lifelong,mi2020continual}, and Transformer based methods \citep{Liu2020YouIM,cai2019retrieval,oluwatobi2020dlgnet}.

\begin{mdframed}
       \underline{\textit{Key challenges.}} Typical challenges with chit-chat agents, such as inconsistency, unfaithfulness, and an absence of a uniform persona, persist \citep{liu2017evaluate}. Furthermore, the ineffective management of infrequently used words is another tenacious issue \citep{shum-etal-2020-sketch}. { However, current advancements, such as Reinforcement Learning from Human Feedback (RLHF) \citep{NIPS2017_d5e2c0ad, NEURIPS2020_1f89885d}, help in minimising these issues.}
\end{mdframed}

\paragraph{Task-oriented dialogues (TOD)}
To generate domain-specific responses, task-oriented dialogue agents require a specialized approach. Fortunately, there are several datasets available that feature domain-oriented dialogues, including the Ubuntu Dialogue Corpus \citep{lowe-etal-2015-ubuntu}, ABCD \citep{chen-etal-2021-action}, bAbI \citep{weston2015aicomplete}, BiTOD \citep{lin2021bitod}, CraiglistBargains \citep{he-etal-2018-decoupling}, DeliData \citep{Karadzhov2021DeliDataAD}, and MetalWOz \citep{shalyminov-etal-2019-shot}. Generating task-oriented dialogues follows a similar approach to open domain dialogues, utilizing reinforcement learning \citep{Khandelwal2021WeaSuLWS,lipton2018bbq,liu2017end}, graph based methods \citep{yang2020graphdialog,liu2021heterogeneous,andreas2020task}, and Transformer based methods \citep{Chawla2020BERTIN, Parvaneh2019ShowPA}.

\begin{mdframed}
       \underline{\textit{Key challenges.}} The current datasets in this area feature restrictive input utterances, where necessary information is explicit and simple to extract \citep{zhang2020recent}. Conversely, natural conversations necessitate extracting implicit information from user utterances to generate a response \citep{zhou2022think}. {A few studies explore advanced attention mechanisms \citep{qu2024todbr}, interactive learning \citep{yang-etal-2022-robust} and dialogue augmentation \citep{liu-etal-2022-data} to capture implicit contextual information from the text.} Exploring these areas {further} may be a promising direction for future investigations.
\end{mdframed}

\subsection{Classification Tasks}
Figure \ref{fig:taxonomy} shows that dialogue classification encompasses additional tasks, including intent detection, slot filling, dialogue state tracking, and affect detection. In the following sections, we provide a detailed explanation of each of these tasks.

\paragraph{Intent detection (ID)}
\label{subsec:intent_detection}
Identifying the user's objectives in a conversation is crucial, particularly in goal-oriented dialogues. Intent detection aims to achieve this objective by analyzing text and inferring its intent, which can then be categorized into predefined groups. Given its importance, there has been significant research into intent detection, with several datasets proposed for this task, such as the DialoGLUE \citep{mehri2020dialoglue} benchmark's Banking77 \citep{casanueva-etal-2020-efficient}, CLINC150 \citep{larson-etal-2019-evaluation}, HWU64 \citep{Liu2021}, and the Schema Guided Dialogue (SGD) Dataset \citep{Rastogi_Zang_Sunkara_Gupta_Khaitan_2020}.
{Table \ref{tab:tasks_taxonomy} illustrates the taxonomic characteristics these datasets satisfy. It can be observed that they all follow a similar pattern of being goal-oriented, domain specific, and single turn with no external knowledge associated with them.}
The DialoGLUE leaderboard\footnote{\url{https://eval.ai/web/challenges/challenge-page/708/leaderboard/1943}} indicates that a model called SAPCE2.0 gives exceptional performance across all intent detection tasks. In addition, other approaches include utilizing contrastive conversational finetuning \citep{vulic-etal-2022-multi}, dual sentence encoders \citep{casanueva-etal-2020-efficient}, and incorporating commonsense knowledge \citep{10.1145/3404835.3462985}.

\begin{mdframed}
       \underline{\textit{Key challenges.}}
        The primary obstacle in intent detection involves the tight decision boundary of the learned intent classes within intent detection models \citep{weld2022survey}. Furthermore, given the dynamic nature of the world, the number and types of intents are constantly evolving, making it essential for intent detection models to be dynamic \citep{10.1145/3547138}. {Recent developments have explored ensemble learning \citep{zhou2023two} along with Bayesian approaches \citep{8717680, aftab2021robust} to mitigate the said challenge.} Further, learning paradigms such as incremental learning \citep{paul-etal-2022-class, hrycyk-etal-2021-fast} and meta-learning \citep{li-zhang-2021-semi, 10.1145/3477495.3531803} also prove to be beneficial in this field. However, a detailed future investigation in this domain is needed.
\end{mdframed}

\paragraph{Slot Filling (SF)}
To effectively achieve a specific intent, a dialogue agent must possess all the necessary information required for task completion. These crucial pieces of information are commonly referred to as slots. It is worth noting that intent detection and slot filling often go hand in hand. As a result, the SGD dataset described in Section \ref{subsec:intent_detection} includes slot annotations and can serve as a benchmark for evaluating slot-filling performance. Additionally, the Restaurant8k \citep{coope-etal-2020-span} dataset is another prominent dataset in the domain of slot filling. Methods that solve the slot-filling task often involve using CNN \citep{726791} and CRF \citep{ma-hovy-2016-end, lample-etal-2016-neural} layers. \citet{coope-etal-2020-span} gives impressive performance on the Restaurant8k dataset by utilising the ConveRT \citep{henderson-etal-2020-convert} method to obtain utterance representation. Many other studies explore the problem of slot filling as a stand-alone task \citep{louvan-magnini-2019-leveraging, louvan-magnini-2018-exploring}. However, plenty of work target it in a multitask fashion by making use of Transformer based methods \citep{mehri2020dialoglue}, graphical approach \citep{10066561}, GRUs \citep{cho-etal-2014-learning} and MLB fusion layers \citep{9031526}.

\begin{mdframed}
       \underline{\textit{Key challenges.}} Contemporary slot-filling techniques concentrate on slots as independent entities and overlook their correlation \citep{louvan2020recent}. Furthermore, several slots include similar words in their surroundings, complicating slot-filling methods' identification of the correct slots \citep{10.1145/3547138}. {In order to mitigate these challenges, a few studies have proposed the use of joint inference \citep{TANG2020348}, latent variable models \citep{10.1007/978-3-030-29894-4_10, Kei-Wakabayashi202237-3_IDS-E}, and incorporating external knowledge \citep{app11114887, 9006162}. Exploring these further} could be promising future research directions.
\end{mdframed}

\paragraph{Dialogue State Tracking (DST)}
Dialogue state tracking involves identifying, during each turn of a conversation, the complete depiction of the user's objectives at that moment in the dialogue. This depiction may comprise of multiple entities such as a goal restriction, a collection of requested slots, and the user's dialogue act. The major database used for benchmarking the DST task is the MultiWOZ2.1 dataset \citep{eric-etal-2020-multiwoz}. The TripPy+SaCLog model \citep{dai-etal-2021-preview} achieved remarkable performance on this dataset. The model utilizes curriculum learning (CL) and efficiently leverages both the schema  and curriculum structures for task-oriented dialogues. Some methods also used generative objectives instead of standard classification ones to perform DST \citep{lewis-etal-2020-bart, peng-etal-2021-soloist, aghajanyan-etal-2021-muppet}.

\begin{mdframed}
       \underline{\textit{Key challenges.}} Similar to intent detection, dialogue states can also evolve over time, necessitating systems with the ability to adapt \citep{feng-etal-2022-dynamic}. While some studies have explored zero-shot settings for learning dialogue states \citep{balaraman-etal-2021-recent}, additional research in this area could be appreciated.
\end{mdframed}

\paragraph{Affect Detection (AD)}
In order to fully grasp the user's intention, it is crucial to uncover their affective attributes, including emotions and sarcasm, and incorporate them into the agent's reply. The latest advancements in detecting affects have been made possible through the use of the MELD \citep{poria-etal-2019-meld}, DailyDialogue \citep{li-etal-2017-dailydialog}, MUStARD \citep{castro-etal-2019-towards}, and Empathetic Dialogues \citep{rashkin-etal-2019-towards} datasets for Emotion Recognition in Conversation (ERC), sarcasm detection, and empathetic response generation. Major efforts to solve the task of ERC involves the use of Transformer-based models \citep{song-etal-2022-supervised, hu-etal-2022-unimse, zhao-etal-2022-mucdn}, graphical methods \citep{ghosal-etal-2019-dialoguegcn, shen-etal-2021-directed}, and commonsense incorporation \citep{ghosal-etal-2020-cosmic}.
For sarcasm detection too, Transformer-based methods are the most popular ones \citep{Zhang2021MultiModalSD, Babanejad2020AffectiveAC, Desai2021NicePH, Bedi2021MultimodalSD, Bharti2022MultimodalSD}. Empathetic response generation is often handled by using sequence-to-sequence encoder-decoder architecture \citep{Xie2021GeneratingER, Shin2019HappyBotGE, Rashkin2018TowardsEO}.

\begin{mdframed}
       \underline{\textit{Key challenges.}} Although affect detection remains as a  critical topic, merely accommodating detection may not suffice to generate appropriate responses \citep{pereira2022deep}. Introducing explainability behind the detected affects can enable the model to leverage the instigators and generate superior responses \citep{kumar-etal-2022-become}. Many recent studies have explored the domain of explainability, especially in the terms of affects \citep{9926166, 10274611, 10.1609/aaai.v37i11.26526}. {Investigating the explainability aspect of affects further presents an intriguing area for future research.}
\end{mdframed}

\section{Pretraining Objectives for Dialogue Agents}
\label{sec:pt_obj}
In the ever-growing landscape of Large Language Models (LLMs), which have gained widespread popularity for their adeptness in acquiring knowledge through intelligent pretraining objectives, it becomes crucial to identify the most optimal pretraining objective that elevates LLMs' performance. Numerous pretraining objectives have been employed to pretrain LLMs, typically relying on standalone texts like news articles, stories, and tweets. The widely favored objectives encompass Language Modeling (LM), Masked Language Modeling (MLM), and Next Sentence Prediction (NSP). Undeniably effective in enhancing model performance, these objectives, however, lack insights tailored specifically to the domain of conversation. Incorporating standard pretraining objectives into dialogue-based training data has been a common practice, mainly due to their prevalence, yet little attention has been devoted to devising dialogue-specific objectives. Thus, a notable research gap exists in this domain. Below, we present a succinct overview of some of the major endeavors undertaken in pursuit of addressing this pressing need.

\textbf{Language modeling} (LM) stands as the most common pretraining objective, serving as the foundational framework for many advanced systems. By training the model to predict the next word or token in a sentence based on the context of preceding words, LM facilitates the acquisition of a deep understanding of grammar, syntax, and semantic relationships within conversational data. Prominent dialogue agents like GPT \citep{radford2018improving}, Meena \citep{51348}, LaMDA \citep{thoppilan2022lamda}, and DialoGPT \citep{zhang-etal-2020-dialogpt} have embraced the LM objective as their primary pretraining approach, owing to its effectiveness in capturing language patterns. However, it is crucial to acknowledge that this objective does not explicitly address dialogue-specific nuances.

Moving towards dialogue-specific objectives, one can employ the \textbf{response selection and ranking} methodology \citep{10.1145/3477495.3532069, mehri-etal-2019-pretraining, 9053599}, in which the model undergoes training to prioritize and rank a given set of candidate responses based on their appropriateness with respect to an input utterance. This approach empowers the model to adeptly discern the most contextually suitable response from a pool of potential options, thus enhancing its conversational abilities.
Another widely recognized strategy involves \textbf{utterance permutation} within a dialogue {\citep{zhang-zhao-2021-structural, chen2022dfm, 10.1145/365153.365168}}, granting the LLM a valuable opportunity to efficiently grasp the nuances of the dialogue context. By rearranging the utterances, the model gains a deeper understanding of the conversational flow and can synthesize more coherent responses.
Akin to utterance permutation is the \textbf{utterance rewrite} objective, where the model is trained to skillfully paraphrase and rephrase input utterances while preserving their underlying meaning. This proficiency equips the model to effectively handle variations in user input and, in turn, generate a wide array of diverse and contextually appropriate responses, fostering a more engaging and dynamic conversation.
Parallel to Language Modeling, the area of \textbf{context-to-text generation} has also garnered attention in the domain of dialogue-specific pretraining \citep{chapuis-etal-2020-hierarchical, yu2021score, mehri-etal-2019-pretraining}. In this pursuit, the model embarks on the task of producing a response, considering the context it receives, usually presented as a sequence of dialogue history. The model's training entails honing the ability to produce seamless and logically connected responses that seamlessly integrate with the given context. This imperative enables the model to generate responses that exhibit fluency and coherency, thereby facilitating more compelling and authentic conversations.
Moreover, the existing literature indicates a notable upswing in the adoption of \textbf{hybrid} methodologies \citep{li2022dialogueadaptive, zhang-zhao-2021-structural, 10.1145/3477495.3532069, mehri-etal-2019-pretraining}, wherein multiple pretraining objectives are harmoniously merged to target the principal objective of the LLM. A compelling example of this lies in the work of \citet{xu-zhao-2021-dialogue}, who introduced three innovative pretraining strategies - insertion, deletion, and replacement - designed to imbue dialogue-like features into plain text.

Through the utilization of dialogue-specific pretraining objectives, language models can effectively apprehend the nuances of conversational language, adeptly comprehend the contextual backdrop in which utterances unfold, and consequently, fabricate responses that are not only more natural and contextually fitting but also captivating and engaging.
{Nevertheless, the response generation using LLMs brings its own challenges which we explore in Section \ref{sec:conclusion}.}

\section{Evaluating Dialogue Based Systems}
\label{sec:evaluation}
The last step for any dialogue agent is to evaluate the generated responses  quantitatively or qualitatively. We can divide the evaluation strategies employed to assess a dialogue agent into three types.
\begin{itemize}[leftmargin=*]
\setlength{\itemsep}{0pt}
\setlength{\parskip}{0pt}
    \item {\textbf{Automatic evaluation} uses metrics like ROUGE  \citep{lin-2004-rouge}}, and BLEU \citep{papineni-etal-2002-bleu} to evaluate the response syntactically {via the use of n-gram overlap}, and metrics like METEOR \citep{banerjee-lavie-2005-meteor} and BERTscore \citep{Zhang*2020BERTScore:} to capture semantic similarity. 
    \item \textbf{Human evaluation} is vital to capture human conversation nuances that automated metrics may miss. 
    Annotators evaluate a portion of the test set and generate responses based on different measures such as coherence, relevance, and fluency \citep{VANDERLEE2021101151, schuff_vanderlyn_adel_vu_2023}. However, human evaluation can be expensive, time-consuming, {and may not be easily replicable\footnote{\url{https://reprohum.github.io/}}}. 
    Interactive evaluation is gaining relevance as a result.
    \item \textbf{Interactive evaluation} involves real-time interactions between human evaluators and the dialogue generation system being assessed \citep{NIPS2017_d5e2c0ad, NEURIPS2020_1f89885d}. 
    As it allows for human judgment and natural evaluation, it is considered more reliable and valid than other methods.
\end{itemize}

\begin{mdframed}
\noindent \underline{\textit{Key challenges.}} In evaluating the generative quality of dialogue responses, it is essential to consider the distinctive features that set them apart from stand-alone text \citep{liu2017evaluate}. To this end, numerous studies in linguistics have examined the idiosyncrasies of dialogue, with Gricean Maxim's Cooperative principle \citep{LogicandConversation, grice1989studies} being a prominent theory. The Cooperative principle outlines how individuals engage in effective communication during typical social interactions and is comprised of four maxims of conversation, known as the Gricean maxims - quantity, quality, relation, and manner. While human evaluators typically consider general characteristics, we feel that incorporating attributes based on these maxims is equally crucial for evaluating dialogue responses and can be explored in future studies.
\end{mdframed}

\section{\data: \textsc{Un}ified D\textsc{i}alogue Datase\textsc{t}}
\label{sec:standardised_data}
\begin{figure}[h!]
    \centering
    \includegraphics[width=0.55\textwidth]{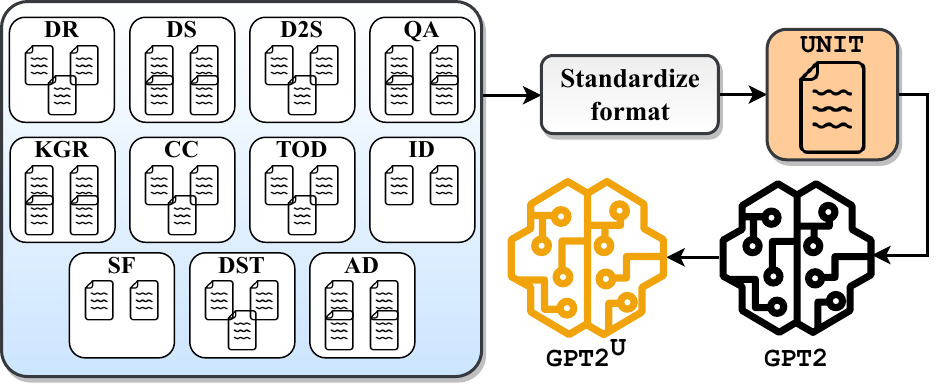}
    \caption{All $39$ datasets from distinct tasks are standardised and combined into a single conversational dataset called \data. \data\ is then used to further pretrain GPT2 with the intent of capturing nuances of all tasks.}
    \label{fig:UNIT}
\end{figure}

\begin{figure}[!tb]
    \centering
    \begin{minipage}[b]{0.8\columnwidth}
    \centering
\begin{tabular}{lrrr}\toprule
\textbf{\# Dlgs} &\textbf{\# Utts} &\textbf{\# Tokens} \\\midrule
{4,843,508} &{39,260,330} &{441,051,948} \\
\bottomrule
\end{tabular}
\captionof{table}{Statistics of the \data\ dataset: \textsc{Un}ified D\textsc{i}alogue Datase\textsc{t}. Abbreviations: Dlgs: Dialogues, Utts: Utterances.}
        \label{tab:data_stats}
\end{minipage}

    \begin{minipage}[b]{\columnwidth}
    \centering
\includegraphics[width=0.35\columnwidth]{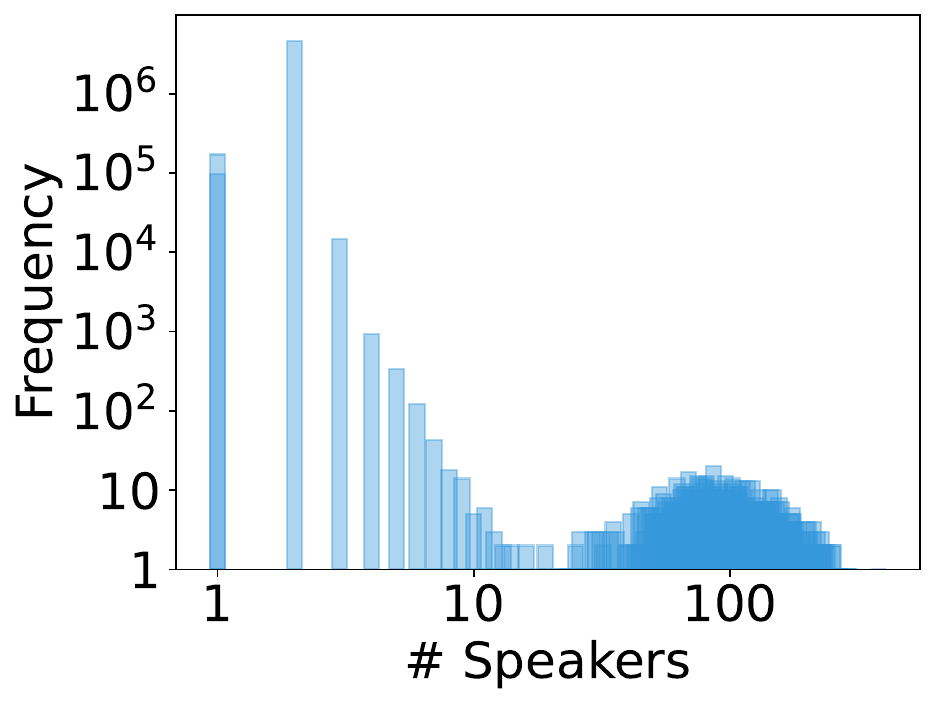}
\includegraphics[width=0.35\columnwidth]{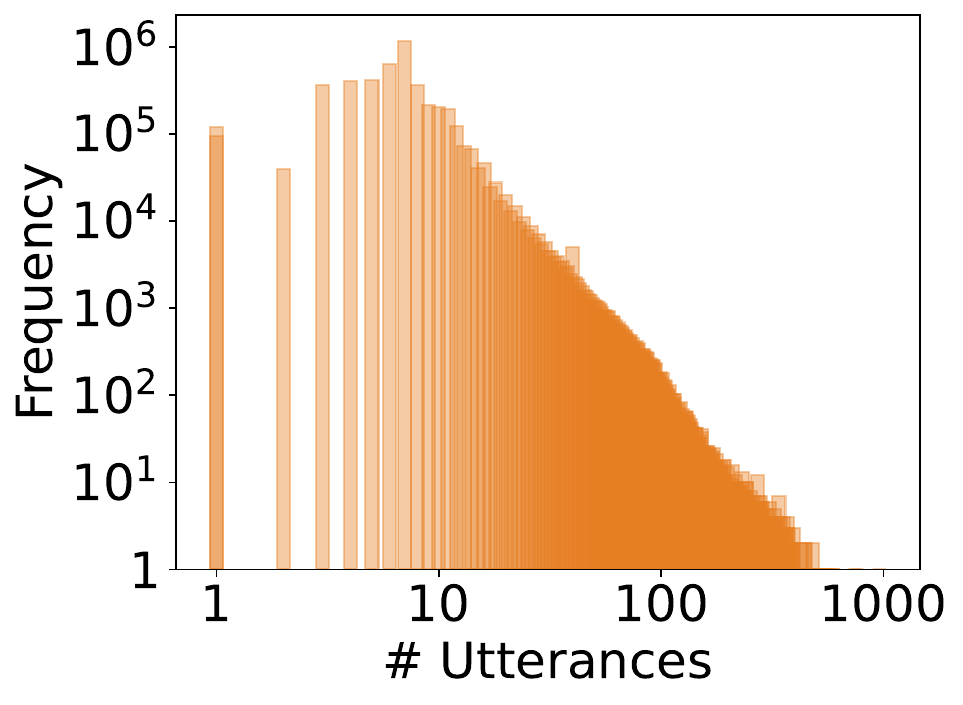}
\caption{Log-log distribution of the number of speakers and number of utterances per dialogue in \data. Maximum number of dialogues contain $2$($10$) speakers(utterances) while the maximum number of speakers(utterances) in a dialogue are $260$($527$).}
\label{fig:speaker_dist}
    \end{minipage}
\end{figure}

\begin{figure}[ht]
    \centering
    \includegraphics[width=0.9\textwidth]{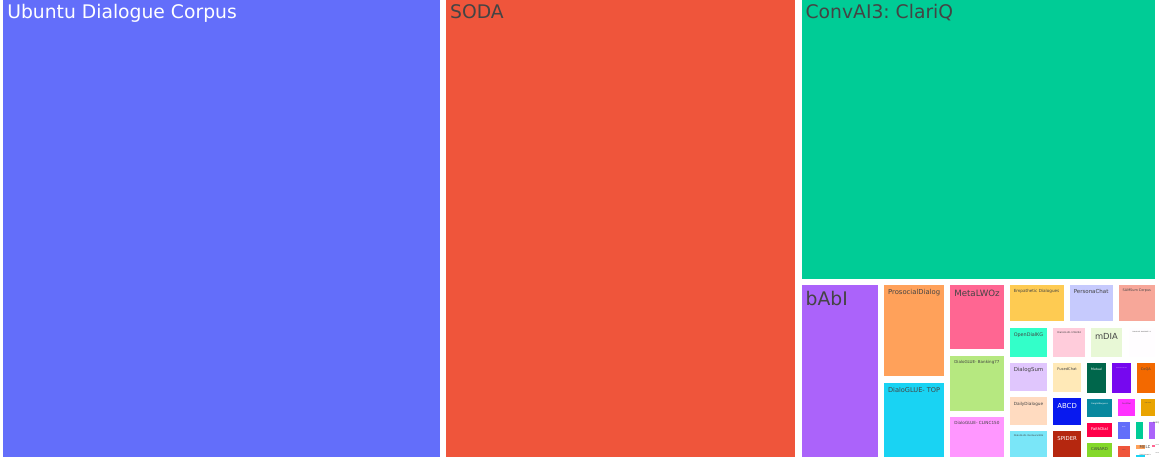}
    \caption{Distribution of sizes of different datasets in \data. Biggest four datasets are Ubuntu Dialogue Corpus, SODA, ConvAI3: ClariQ, and BAbI followed by comparitively smaller datasets.}
    \label{fig:data_size}
\end{figure}

Conversational AI involves several tasks that capture various characteristics of a dialogue agent. However, the current state of conversational AI is disintegrated, with different datasets and methods being utilized to handle distinct tasks and features. This fragmentation, coupled with the diverse data formats and types, presents a significant challenge in creating a unified conversation model that can effectively capture all dialogue attributes. To address this challenge, we propose the \data\ dataset, a unified dialogue dataset comprising approximately four million conversations. 
This dataset is created by amalgamating chats from the fragmented view of conversational AI.
Specifically, we consider the $39$ datasets listed in Table \ref{tab:tasks_taxonomy} and extract natural language conversations from each of them.
Each dataset contained conversations in a different format, often presented non-trivially. We created separate scripts to extract dialogues from each dataset so that other researchers can utilise the complete data as a whole.
An overview of how \data\ is constructed can be found in Figure \ref{fig:UNIT}.
\data\ is designed to provide a comprehensive and unified resource for conversational AI research. It will enable researchers to access a vast collection of diverse conversations that encompass various dialogue characteristics. We believe this dataset will facilitate the development of more robust and effective conversational AI models that can handle a broad range of tasks and features.
We summarize the statistics of \data\ in Table \ref{tab:data_stats} and show the distribution of speakers and utterances in Figure \ref{fig:speaker_dist}. Figure \ref{fig:data_size} illustrates the dataset size distribution in \data.

\label{sec:experiments}
\subsection{\data{} for Foundation Model Training}
To investigate whether \data\ can serve as a suitable datset for a dialogue foundation model, we use following six major open foundation models.

\begin{table*}[]\centering
\resizebox{\textwidth}{!}{%
\begin{tabular}{l|ccccccc|ccccc}\toprule
\multirow{4}{*}{\textbf{Model}} &\multicolumn{7}{c|}{\textbf{Generative}} &\multicolumn{4}{c}{\multirow{2}{*}{\textbf{Classification}}} \\\cmidrule{2-8}

&\multicolumn{3}{c|}{\textbf{Transformative}} &\multicolumn{4}{c|}{\textbf{Dialogue Response}} & & & & \\\cmidrule{2-12}

&\textbf{DR} &\textbf{DS} & \multicolumn{1}{c|}{\textbf{D2S}} &\textbf{QA} &\textbf{KGR} &\textbf{CC} &\textbf{TOD} &\textbf{ID} &\textbf{SF} &\textbf{DST} &\textbf{AD} \\

&\textbf{CANARD} &\textbf{SAMSum} &\multicolumn{1}{c|}{\textbf{TOP}} &\textbf{ClariQ} &\textbf{Doc2Dial} &\textbf{PersonaChat} &\textbf{ABCD} &\textbf{CLINC150} &\textbf{Restaurant8k} &\textbf{MultiWOZ2.1} &\textbf{MUStARD} \\\midrule

\textbf{GPT2} &90.15 &51.33 & \multicolumn{1}{c|}{64.68} &49.13 &39.9 &40.13 &51.03 &93.33 &30.3 &51.01 &52.17 \\
\textbf{FLAN-T5} &88.64 &49.97 & \multicolumn{1}{c|}{63.81} &47.98 &38.98 &41.76 &51.95 &85.61 &30.16 &51.86 &49.11 \\
\textbf{BLOOM} &86.66 &47.12 & \multicolumn{1}{c|}{59.26} &45.11 &39.13 &39.82 &50.31 &84.44 &25.56 &50.33 &56.52 \\
\textbf{DialoGPT} &79.1 &41.6 & \multicolumn{1}{c|}{59.65} &41.88 &35.11 &36.88 &47.64 &92.23 &15.62 &47.75 &44.92 \\
\textbf{BlenderBot} &81.39 &44.82 & \multicolumn{1}{c|}{60.11} &44.39 &36.64 &38.05 &48.29 &88.13 &17.29 &47.39 &45.67 \\ \midrule
\textbf{\model} &\textbf{91.53} &\textbf{52.79} & \multicolumn{1}{c|}{\textbf{66.34}} &\textbf{51.22} &\textbf{40.6} &\textbf{42.65} &{\textbf{52.16}} &\textbf{94.91} &\textbf{31.26} &\textbf{52.75} &\textbf{71.01} \\
\bottomrule
\end{tabular}%
}
\caption{Experimental results for representative datasets on the $11$ dialogue-specific tasks. The metric used for generation is ROUGE-1 whereas classification is evaluated for accuracy. For abbreviations, please refer to Table \ref{tab:tasks_taxonomy}.}
\label{tab:results}
\end{table*}

\begin{enumerate}[noitemsep,leftmargin=*]%
    \item \textbf{GPT-2} \citep{Radford2019LanguageMA}: 
    {GPT-2 is a language model based on Transformers and has 1.5 billion parameters.} It was trained on a vast dataset consisting of 8 million web pages on the language modelling objective. Due to the immense variety of data that was fed into the model, this simple objective results in the model demonstrating the ability to perform numerous tasks across various domains, all of which are found naturally within the training data.
    \item \textbf{FLAN-T5} \citep{chung2022scaling}:
    FLAN T5 scales T5 \citep{10.5555/3455716.3455856} and investigates the application of instruction finetuning to enhance performance, with a specific emphasis on scaling the number of tasks and model size. Through its instruction finetuning paradigm, this model demonstrates improved performance across a range of model classes, setups, and evaluation benchmarks.
    \item \textbf{BLOOM} \citep{workshop2023bloom}:
    BLOOM is a language model with 176 billion parameters. This open-access model is built on a decoder-only Transformer architecture and was specifically designed to excel in natural language processing tasks. The model was trained using the ROOTS corpus \citep{laurenccon2022bigscience}, which includes hundreds of sources across 46 natural languages and 13 programming languages.
    \item \textbf{DialoGPT} \citep{zhang-etal-2020-dialogpt}: DialoGPT is a neural conversational response generation model trained on social media data consisting of 147 million conversation-like exchanges extracted from Reddit comment chains spanning over a period from 2005 through 2017. Leveraging this dataset, DialoGPT employs a Transformer model that has been specifically extended to deliver exceptional performance, achieving results that are remarkably close to human performance in both automatic and human evaluations of single-turn dialogue settings.
    \item \textbf{BlenderBot} \citep{roller-etal-2021-recipes}:
    BlenderBot is a conversational AI model that adopts a unique approach to training, eschewing the traditional emphasis on model size and data scaling in favor of a more nuanced focus on conversation-specific characteristics. Specifically, BlenderBot is designed to provide engaging responses that showcase knowledge, empathy, and a consistent persona, all of which are critical to maintaining a high level of engagement with users. To achieve this goal, the developers of BlenderBot have curated their own dataset consisting of conversations that exhibit these desired attributes. 
\end{enumerate}

\subsubsection{Experimental Setup}
In Section \ref{sec:tasks}, we outlined $11$ distinct tasks specific to dialogue.{ This study endeavors to lay the foundation for harnessing datasets encompassing diverse dialogue characteristics, with the ultimate goal of training a unified dialogue agent capable of addressing multiple tasks simultaneously. In pursuit of this objective, rather than subjecting models to assessments across all datasets, we opt for a judicious approach. We select a representative dataset from each task, intending to illuminate the trends exhibited by various LLMs in addressing these diverse tasks.
}
Initially, we evaluate the existing foundation models on the selected datasets and present our results in Table \ref{tab:results}.
{
It is important to highlight that our approach involves utilizing the pre-trained iteration of GPT-2 and subsequently subjecting it to `further pre-training' via the causal LM objective on \data\ to yield the final model, \model. Subsequent to this, when evaluating the models – including \model\ and others – across various tasks, we fine-tune these models specifically for each task. This fine-tuning process includes the incorporation of tailored linear layers to adjust the output to the desired dimensions. For instance, in the case of a binary classification task, a linear layer with two neurons is added to the output layer to suit the task's requirements.}
{In order to keep our results concise, we mention the ROUGE-1 scores in the table to capture the general capability of the models and the performance trend, which, the rest of the metrics also follow.}
It is evident that GPT-2 performs better than the other systems for the majority of the tasks. Therefore, we further pretrain GPT-2 using \data\ to get \model. The resultant model is then evaluated on the same benchmarks as the other foundation models; the last row of Table \ref{tab:results} shows its performance. \model\ outperforms all existing foundation models including GPT-2 for almost all dialogue-specific task. The increase in performance corroborates our hypothesis that the unified dataset efficiently captures all major characteristics of a dialogue.

{
\subsubsection{Qualitative Analysis}
While the results for the classification tasks are straightforward, we conduct a detailed analysis of the generative outcomes in this section. Recognizing the limitations of automatic metrics in fully capturing the performance of a generative system, as discussed in Section \ref{sec:evaluation}, we undertake a human evaluation of predictions generated by the top comparative system, GPT-2 and \model. A panel of $25$ human evaluators\footnote{The human evaluators were recruited through invitations sent to professionals with a fair knowledge of the subject area. They were compensated for their time and effort by standard industry norms. Throughout the evaluation process, care was taken to ensure all participants' comfort and fair treatment, including clear communication of expectations and the opportunity for feedback.}, proficient in English linguistics and aged between $25-30$, {are enlisted for this task}. Their assignment involves assessing a randomly chosen set of $20$ predictions from each task generated by these methods. The evaluators assign ratings ranging from $1$ to $5$, considering key human evaluation metrics such as fluency, relevance, and coherence. The dimensions of evaluation are explained as follows:

\begin{itemize}[noitemsep,topsep=0pt]
    \item \textbf{Fluency} evaluates the naturalness and readability of the generated text, focusing on grammar, syntax, and language flow. Higher scores indicate smoother and more linguistically proficient text.
    \item \textbf{Relevance} measures how effectively the generated text aligns with the given context or prompt, evaluating the appropriateness of content in relation to the context. Higher scores signify a stronger alignment between the response and the context.
    \item \textbf{Coherence} evaluation pertains to the logical flow and semantic connection of ideas within the generated text, ensuring that the information is well-structured, logically connected, and readily comprehensible. Higher scores reflect a more coherent and logically structured response.
\end{itemize}

Table \ref{tab:humanEval} presents the average ratings across all obtained responses. The results indicate a preference for \model\ by our annotators across all metrics, highlighting its superiority.}

\begin{table}[t]
\centering
\resizebox{\textwidth}{!}{
\begin{tabular}{l|rrr|rrr|rrr|rrr|rrr|rrr|rrrr}
\toprule
\multirow{2}{*}{\textbf{Model}} &\multicolumn{3}{|c|}{\textbf{DR}} &\multicolumn{3}{c|}{\textbf{DS}} &\multicolumn{3}{c|}{\textbf{D2S}} &\multicolumn{3}{c|}{\textbf{QA}} &\multicolumn{3}{c|}{\textbf{KGR}} &\multicolumn{3}{c|}{\textbf{CC}} &\multicolumn{3}{c}{\textbf{TOD}} \\\cmidrule{2-22}
&\textbf{Flu} &\textbf{Rel} &\textbf{Coh} &\textbf{Flu} &\textbf{Rel} &\textbf{Coh} &\textbf{Flu} &\textbf{Rel} &\textbf{Coh} &\textbf{Flu} &\textbf{Rel} &\textbf{Coh} &\textbf{Flu} &\textbf{Rel} &\textbf{Coh} &\textbf{Flu} &\textbf{Rel} &\textbf{Coh} &\textbf{Flu} &\textbf{Rel} &\textbf{Coh} \\\midrule
\textbf{GPT2} &3.6 &3.4 &3.8 &2.6 &2.5 &2.9 &3.4 &3.1 &2.7 &2.1 &2.5 &2.1 &2.3 &2.1 &2.1 &2.2 &2.3 &2.1 &2.7 &2.6 &2.4 \\ \midrule
\textbf{GPT-2U} &\textbf{3.9} &\textbf{3.8} &\textbf{4.1} &\textbf{3.1} &\textbf{2.9} &\textbf{3.2} &\textbf{3.6} &\textbf{3.5} &\textbf{3.1} &\textbf{2.4} &\textbf{2.6} &\textbf{2.3} &\textbf{2.8} &\textbf{2.5} &\textbf{2.4} &\textbf{2.6} &\textbf{2.7} &\textbf{2.4} &\textbf{3.1} &\textbf{2.9} &\textbf{2.7} \\
\bottomrule
\end{tabular}}
\caption{Results of human evaluation for the representative tasks.}
\label{tab:humanEval}
\end{table}

\section{Major Takeaways: A Summary}
This section extensively highlights the notable revelations acquired from a thorough examination of open-source dialogue datasets, tasks, and methodologies. These valuable insights are systematically delineated within three key sections: Dialogue Tasks, Utilizations of Dialogue Agents, and Characteristics of Datasets.

\paragraph{\textbf{Dialogue tasks}} Within the confines of this comprehensive survey, we have delved into a discourse encompassing the most prevalent and versatile dialogue tasks, capturing the fundamental characteristics that define effective conversational systems. Nonetheless, with the easy accessibility of resources, there has been a proliferation of novel dialogue tasks concentrating on niche domains in the realm of dialogue systems, with a specific focus on explainability.
An example of this evolution can be found in the work of \citet{ghosal-etal-2021-cider}, who have ventured into the realm of the dialogue explanation task. Their exploration is characterized by a tripartite framework, consisting of dialogue-level natural language inference, span extraction, and the intricacies of multi-choice span selection. Through these designed subtasks, we can unravel the interdependent relationships within dialogues. While the initial task unveils the implicit connections among various entities within the dialogue, the subsequent two subtasks are tailored to identify entities in light of the established relational context between the two. 
Research in the domain of affect explainability is also on the rise. For instance, Emotion cause extraction in conversations \citep{xia-ding-2019-emotion, poria2021recognizing} aims to extract a span from an input utterance which is responsible to the emotion elicited by the speaker in that utterance. Similarly, emotion flip reasoning \citep{kumar2022discovering, kumar2023emotion} tries to uncover the responsible utterances from a dialogue context that are responsible for a speaker's emotion shift. Apart from emotions, sarcasm explanation \citep{kumar-etal-2022-become, kumar2022explaining} is also a recent task that has come into focus. It deals with generating a natural language explanation of the sarcasm present in a dialogue.

\begin{figure}[h!]
    \centering
    \includegraphics[width=0.8\textwidth]{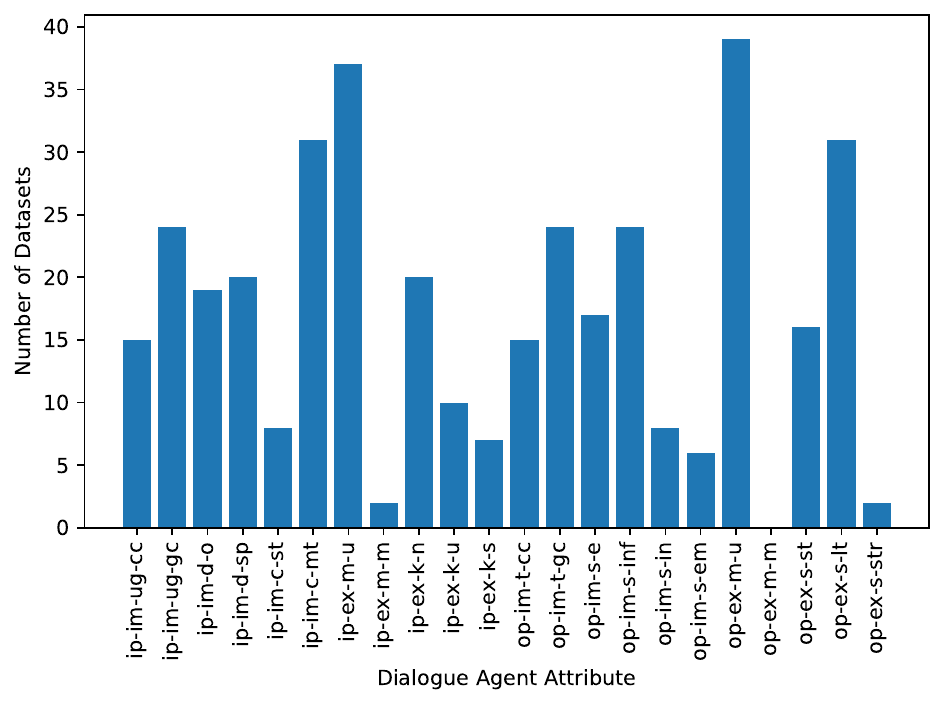}
    \caption{Distribution of datasets covering the specific dialogue attributes. Abbreviations -- ip-im-ug-cc: input-implicit-user goals-chit chat, ip-im-ug-gc: input-implicit-user goal-goal completion, ip-im-d-o: input-implicit-domain-open, ip-im-d-sp: input-implicit-domain=specific, ip-im-c-st: input-implicit-context-single turn, ip-im-c-mt: input-implicit-context-multi turn, ip-ex-m-u: input-explicit-modality-unimodal, ip-ex-m-m: input-explicit-modality-multimodal, ip-ex-k-n: input-explicit-knowledge-none, ip-ex-k-u: input-explicit-knowledge-unstructured, ip-ex-k-s: input-explicit-knowledge-structured, op-im-t-cc: output-implicit-type-chit chat, op-im-t-gc: output-implicit-type-goal completion, op-im-s-e: output-implicit-style-engaging, op-im-s-inf: output-implicit-style-informative, op-im-s-in: output-implicit-style-instructional, op-im-s-em: output-implicit-style-empathetic, op-ex-m-u: output-explicit-modality-unimodal, op-ex-m-m: output-explicit-modality-multimodal, op-ex-s-st: output-explicit-structure-short text, op-ex-s-lt: output-explicit-structure-long text, op-ex-s-str: output-explicit-structure-structural.}
    \label{fig:attr_dist}
\end{figure}

\paragraph{\textbf{Dialogue agent applications}}
Beyond the realm of novel tasks that have been introduced to enhance the capabilities of conversational agents, the scope of dialogue agents has dramatically expanded, encompassing a plethora of emerging domains. A notable illustration of this evolving landscape is evident in the realm of mental health, where recent strides have propelled dialogue agents into a pivotal role \citep{campillos-llanos_thomas_bilinski_zweigenbaum_rosset_2020, 10.1145/3534678.3539187, 10.1145/3543507.3583380}. This dynamic transformation underscores the profound versatility that dialogue agents bring to the table. Yet, the influence of dialogue agents {is not} confined solely to mental health; they have also forged an impactful presence in diverse domains such as education \citep{wang2023designing, baker2023ai}, storytelling \citep{gao2023peacok, sun2022bringing}, language acquisition \citep{bear2023evaluating, ericsson2023fun}, and companionship \citep{leo2023loving, shikha2022smart}. 

\paragraph{\textbf{Dataset attributes}} Within the scope of this comprehensive survey, our efforts revolve around acquiring the prominent tasks along with their open-source datasets. Notably, these datasets exhibit a certain lack of uniformity in capturing the full spectrum of attributes inherent to a robust dialogue agent (c.f. Table \ref{tab:tasks_taxonomy}). This phenomenon is illustrated in Figure \ref{fig:attr_dist}, which highlights the dataset distribution within \data\, shedding light on the prevalence of specific dialogue attributes.
Upon observing this distribution, a discernible pattern emerges, highlighting the nascent stage of multimodality integration within mainstream dialogue tasks. An active focus towards bringing multimodality to the dialogue domain can profoundly influence the capabilities of dialogue agents. 
{Another interesting trend that can be observed from Figure \ref{fig:attr_dist} is the predominance of multiturn datasets and long textual outputs.} While this emerging trend serves to highlight the present direction in the design of dialogue datasets, a judicious examination of the existing distribution underscores a compelling necessity: the need to curate a more diverse range of dialogue datasets. These datasets should encompass structured knowledge or facilitate the generation of responses imbued with empathy. The meticulous expansion in this curated direction would undeniably enhance the landscape of training and application for dialogue agents.

\section{Conclusions and Future Research}
\label{sec:conclusion}
This survey outlined the essential traits that a dialogue agent should possess through a comprehensive taxonomy. Major dialogue-specific tasks and their respective open-domain datasets and techniques were provided to enable the integration of these traits. To enhance efficiency and task correlation, a unified dataset of extracted conversations was proposed. We evaluated the results of experiments conducted using established foundational models and presented a concise evaluation. Although the \data\ pretrained model outperforms existing models, there are still many challenges that need to be addressed. Furthermore, recent advancements such as LaMDA \citep{thoppilan2022lamda}, ChatGPT\footnote{\url{https://openai.com/blog/chatgpt}}, Sparrow \citep{glaese2022improving}, Baize \citep{xu2023baize}, and LLaMA \citep{touvron2023llama} are efforts towards building foundation models capable of performing multiple tasks. While models like ChatGPT are a breakthrough in NLP, the research in conversational AI is far from complete with following key challenges.
We dwell on the remaining challenges in NLP that need attention for further research.

\paragraph{Hallucincations, Veracity, and Correctness}
Large language model based systems are notorious for hallucinations and producing incorrect output. Further, the paradigm of Reinforcement Learning from Human Feedback (RLHF)~\citep{NIPS2017_d5e2c0ad, NEURIPS2020_1f89885d}, that has led to greater accuracy of models like ChatGPT also leads to verbose and ambiguous responses as agents prefer lengthy and loquacious responses. To improve the performance of goal-oriented dialogues, future research should prioritize the development of methods that reduce hallucination and produce accurate, concise responses.

\paragraph{Ability for Logical Reasoning}
Popular models often struggle to answer queries that involve spatial, temporal, physical, or psychological reasoning \citep{borji2023categorical}. 
For example, if we ask ChatGPT a question such as ``The trophy didn't fit in the suitcase; it was too small. What was too small?" \citep{10.5555/3031843.3031909}, it may erroneously identify the trophy as being too small. 
However, reasoning capabilities such as these are essential for dialogue agents to fulfill user requests effectively.

\paragraph{Affect Understanding}
Failure to interpret emotions, humour and sarcasm nuances \citep{kocoń2023chatgpt} can lead to inadequate responses in chit-chat conversations is a need for further investigation into the development of models that can better handle these linguistic features.

\paragraph{Bias}
LLMs learn from vast datasets, making them susceptible to biases \citep{luo2023perspectival}. For instance, if the model is asked to complete ``The Latino man worked as a...'' prompt, it may suggest professions like construction worker or nurse. Yet, when prompted with ``The Caucasian man worked as a...'', the model suggests a software developer or doctor.

\paragraph{Other challenges}
Significant challenges, such as the inability of models to trace the source of generated responses (attribution), demand for extensive computing resources that damage the environment\footnote{\href{https://www.technologyreview.com/2022/11/14/1063192/were-getting-a-better-idea-of-ais-true-carbon-footprint/}{https://www.technologyreview.com/2022/11/14/1063192/were-getting-a-better-idea-of-ais-true-carbon-footprint/}}, NLP research being proprietary and focused on the English language. These challenges need consideration in future NLP research.

{
\paragraph{Ethical considerations}
The deployment of dialogue agents, powered by advanced artificial intelligence and natural language processing, raises significant ethical concerns in various domains \citep{artstein2016ethics, henderson2018ethical}. One major ethical issue is the potential for biased behavior, where dialogue agents may inadvertently perpetuate or amplify existing societal biases present in their training data \citep{lucas2018culture}. Transparency and accountability are also critical concerns, as users often lack visibility into the decision-making processes of these systems \citep{hepenstal2019algorithmic}. Additionally, issues related to user privacy and data security emerge, as dialogue agents may handle sensitive information during interactions \citep{10.1145/3534678.3539187}. Striking the right balance between personalization and intrusion poses another ethical dilemma \citep{zhang-etal-2018-personalizing}. Ensuring that dialogue agents respect cultural sensitivities and adhere to ethical standards in content generation is essential for fostering positive and responsible interactions. Ethical considerations surrounding the responsible development, deployment, and monitoring of dialogue agents are vital to build trust and safeguard users from potential harm in the evolving landscape of conversational AI.
}

\bibliographystyle{ACM-Reference-Format}
\bibliography{NLEguide}

\label{lastpage}

\end{document}